\documentclass[nohyperref]{article}

\usepackage{microtype}
\usepackage{graphicx}
\usepackage{subfigure}
\usepackage{booktabs} 

\usepackage{hyperref}

\usepackage[accepted]{icml2022}

\usepackage{amsmath}
\usepackage{amssymb}
\usepackage{mathtools}
\usepackage{amsthm}

\usepackage[capitalize,noabbrev]{cleveref}

\theoremstyle{plain}
\newtheorem{theorem}{Theorem}[section]
\newtheorem{proposition}[theorem]{Proposition}

\newtheorem{corollary}[theorem]{Corollary}
\theoremstyle{definition}

\theoremstyle{remark}
\newtheorem{remark}[theorem]{Remark}

\usepackage[textsize=tiny]{todonotes}

\icmltitlerunning{Exponentially Tilted Gaussian Prior for Variational Autoencoders}

\usepackage{mathtools}
\DeclarePairedDelimiterX{\infdivx}[2]{(}{)}{%
  #1\;\delimsize\|\;#2%
}
\newcommand{\infdiv}{D_{\mathrm{KL}}\infdivx}
\DeclarePairedDelimiter{\norm}{\lVert}{\rVert}

\begin{document}

\twocolumn[
\icmltitle{The Exponentially Tilted Gaussian Prior for Variational Autoencoders}




\begin{icmlauthorlist}
\icmlauthor{Griffin Floto}{engg}
\icmlauthor{Stefan C. Kremer}{comp}
\icmlauthor{Mihai Nica}{math}
\end{icmlauthorlist}

\icmlcorrespondingauthor{Griffin Floto}{gfloto@mail.uoguelph.ca}
\icmlaffiliation{engg}{School of Engineering, University of Guelph}
\icmlaffiliation{comp}{School of Computer Science, University of Guelph}
\icmlaffiliation{math}{Department of Mathematics \& Statistics, University of Guelph}

\icmlkeywords{Variational Autoencoders, Out of Distribution Detection}

\vskip 0.3in
]



\printAffiliationsAndNotice{\icmlEqualContribution} 

\begin{abstract}
An important property for deep neural networks is the ability to perform robust out-of-distribution detection on previously unseen data. This property is essential for safety purposes when deploying models for real world applications. Recent studies show that probabilistic generative models can perform poorly on this task, which is surprising given that they seek to estimate the likelihood of training data. To alleviate this issue, we propose the exponentially tilted Gaussian prior distribution for the Variational Autoencoder (VAE) which pulls points onto the surface of a hyper-sphere in latent space. This achieves state-of-the art results on the area under the curve-receiver operator characteristics metric using just the log-likelihood that the VAE naturally assigns. Because this prior is a simple modification of the traditional VAE prior, it is faster and easier to implement than competitive methods. 

\end{abstract}

\section{Introduction}

For deep learning models to be safely deployed, they must be able to make reliable and accurate predictions when faced with data that is outside the training distribution. Data of this type is referred to as out-of-distribution (OOD), and highlights serious safety concerns if deep learning models cannot perform as intended when receiving OOD data. For example, in the case of medical diagnosis a serious failure case would occur if a given sample is OOD and a model confidently returns a negative diagnosis. Problems of this type can generally be described as robustness to distributional shift, which is currently an open problem and is crucial to the development of effective  AI systems \citep{Amodei}. A method of addressing distributional shift is to use an OOD detection model to determine if a sample is from the training distribution before making a prediction, and to decline any samples that are identified as OOD. A natural dichotomy of OOD detection approaches are between those which are supervised and unsupervised. The benefit of unsupervised methods is that they are more practical to use given that labelling data is very expensive. This type of solution will be investigated in our study. 

\begin{figure}[ht]
\begin{center}
\centerline{\includegraphics[width=\columnwidth]{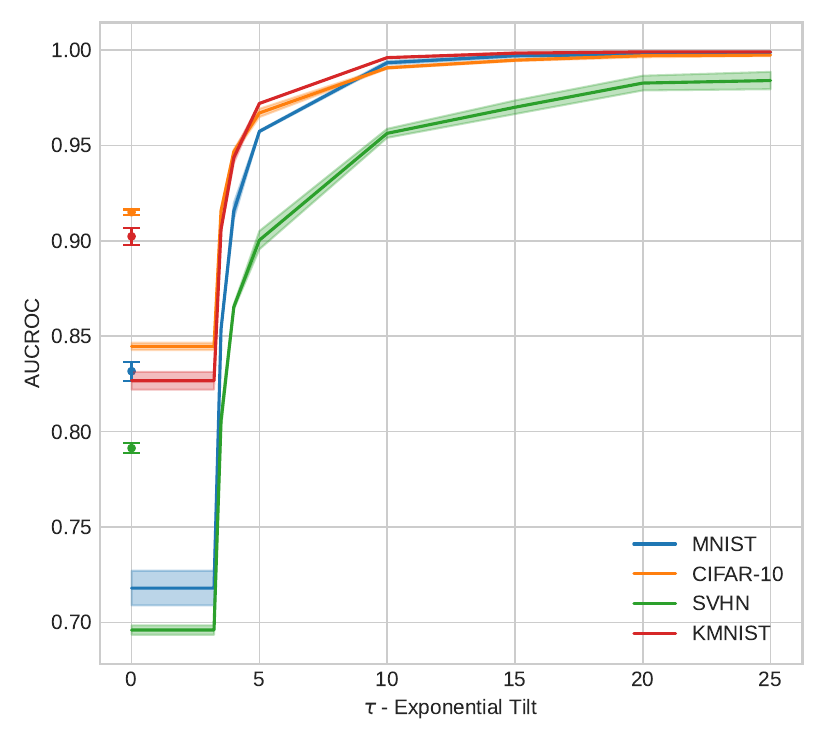}}
\caption{Models trained on the Fashion-MNIST dataset with a latent dimension $d_z = 10$ are evaluated on 4 different OOD datasets using the AUCROC metric. Lines denote models trained with the proposed tilted prior under an increasing $\tau$ parameter, while scatter points on the $y$-axis denote models trained with a standard Gaussian prior. Standard deviations for $N=3$ trials are plotted for both models. We observe that the tilted prior has degraded performance with small $\tau$ due to the simplified covariance matrix assumptions, while performance rapidly improves and surpasses the standard Gaussian prior as $\tau$ increases.}
\label{cover}
\end{center}
\end{figure}

Deep probabilistic generative models such as variational autoencoders (VAEs) \citep{Kingma2}, normalizing flows \citep{Kingma3} and auto-regressive models \citep{vandenoord2} are appealing candidates for methods that could successfully detect OOD samples in an unsupervised manner given that they approximate the likelihood of training data. Despite this, it has been discovered that these models perform surprisingly poorly on OOD data \citep{Nalisnick1, HChoi}. To remedy this, numerous new methods have been developed, however to the best of our knowledge no general solution exists. Many of these studies propose new scores rather than using the log-likelihood which is a natural statistic of the model. Intriguingly, there appears to be an important difference between VAEs and other generative models with respect to OOD detection. A notable study by \citet{Xiao} found that many of the solutions to the OOD problem that were effective on generative models like normalizing flows, performed significantly worse on VAEs. 

Some works have shown that replacing the standard Gaussian prior typically used for VAEs with an alternative prior results in better OOD detection performance, without the introduction of new scores \citep{Havtorn, Maaloe2}. Following this line of work, our proposed method uses an exponentially tilted version of the standard Gaussian prior for VAEs. The benefit of this method for OOD detection is that additional training data and augmentation are not required, and runtime is not affected. We compare existing methods for OOD detection with VAEs and demonstrate that our proposed method matches or improves upon the state-of-the-art in a number of tests.

\section{Method}
\subsection{Review of Variational Bound on the Marginal Likelihood with a Gaussian Prior}
Consider a dataset $X = \{ x^i \}^N_{i=1}$ that consists of $N$ i.i.d. samples of some variable $x^i \in \mathbb{R}^{d_x}$. We assume that the data is produced by a random process which is conditioned on a latent unobserved variable $z \in \mathbb{R}^{d_z}$. The data generation process consists of first sampling the prior distribution $p_{\theta^*}\left (z\right)$, then a value $x^i$ is produced by the generator or decoder $p_{\theta^*}\left (x\vert z\right)$. Both the true parameters $\theta^*$ and the value of the latent variables $z^i$ are unknown. We would like to maximize the marginal log-likelihood of the data $\sum^{N}_{\stackrel{i =1}{}} \textrm{log } p_{\theta}(x^i)$, however the integral over the latent variables is intractable. The variational lower bound is commonly used to deal with this problem by introducing an inference model or encoder $q_{\phi}\left (z\vert x\right)$ to approximate the true posterior $p_{\theta}\left(z\vert x\right)$. Given an encoder model, the variational lower bound of the marginal likelihood can be written as

\begin{align}
\label{vbound}
\log p\left(x^i\right) \geq\textrm{ } &\mathbb{E}_{q_{\phi}(z \vert x^i)} \left[\log p_{\theta}\left(x^i \vert z\right)\right]\notag\\
&-\infdiv{q_{\phi}(z\vert x)}{p_{\theta}(z)},
\end{align}

where $D_{\mathrm{KL}}\left(\cdot\right)$ is the Kullback-Leibler divergence (KLD). The KLD term in the equation above is commonly interpreted as fitting the aggregated posterior $q_{\phi}\left (z\vert x\right)$ to a pre-determined prior $p_{\theta}\left (z\right)$. Typically, the standard Gaussian distribution $\mathcal{N}\left(0,I\right)$ is used as the prior. By assuming an encoder distribution of the form $\mathcal{N}\left(\mu,\textrm{diag}(\sigma\right))$ where $\sigma \in \mathbb{R}^{d_z}$ the model then has a KLD of $\infdiv{\mathcal{N}\left(\mu,\textrm{diag}(\sigma)\right)}{\mathcal{N}\left(0,I\right)}$ which can be evaluated explicitly as

\begin{align}
&\infdiv{\mathcal{N}\left(\mu,\textrm{diag}(\sigma)\right)}{\mathcal{N}\left(0,I\right)} \notag \\
=& \frac{1}{2}\sum^{d_z}_{\stackrel{j =1}{}} \left(\sigma_j^2 +\mu_j^2 -1 -\log \sigma_j^2 \right).
\end{align}

This simple formula makes the lower bound of \eqref{vbound} explicit, which can then be optimized using stochastic gradient descent. Gaussian noise is added to each point during training so that the latent point $z^i$ assigned to the data point $x^i$ is $z^i \sim \mathcal{N}(\mu(x^i), \sigma(x^i))$. This is sometimes called the  ``reparameterization trick" \citep{Kingma2}.



\subsection{Exponentially Tilted Gaussian Prior}
The primary technical contribution of this paper is the proposal of an alternative prior distribution, namely the exponentially tilted Gaussian prior (or simply the ``tilted prior''). The ordinary Gaussian prior has maximum density at the origin, which means KLD term in the log-likelihood bound \eqref{vbound} forces all latent points into the same location, $\mu_i = 0$ and $\sigma_i = 1$. This can lead to ``crowding'' around the origin, making it difficult to differentiate between encoded data samples \citep{Hoffman, Alemi}. 

\begin{figure}[ht]
\begin{center}
\centerline{\includegraphics[width=\columnwidth]{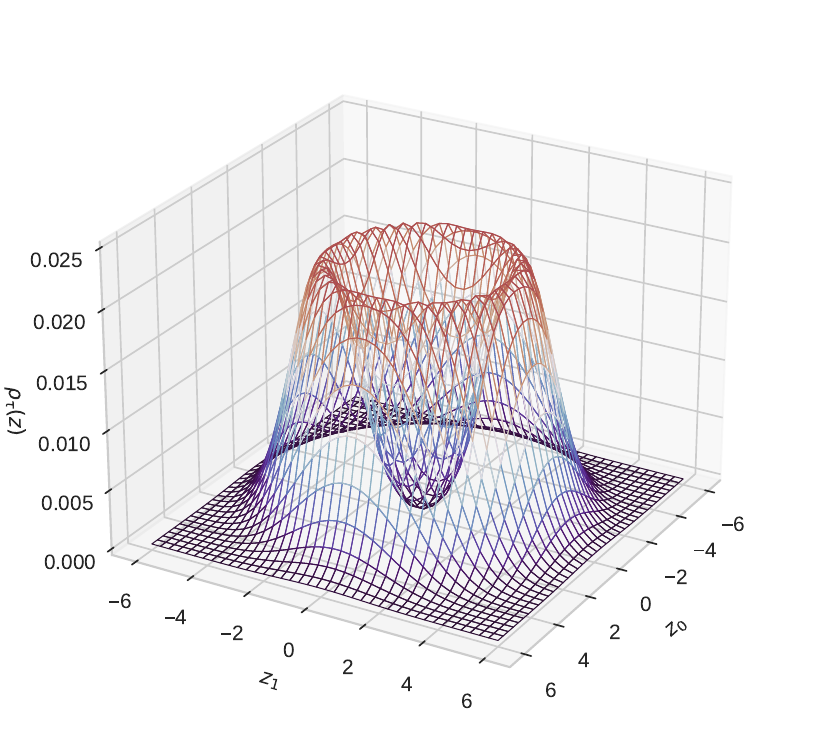}}
\caption{Probability density function $\rho_\tau(z)$ of the exponentially tilted Gaussian distribution $\mathcal{N}(0,I)$ for $\tau = 3, d_z = 2$ }
\label{distribution}
\end{center}
\end{figure}

In contrast to the standard Gaussian, the titled prior does not have maximum density at a single point. Instead, the maximum density occurs at all the points on the hypershere of radius $\tau$ as illustrated in Figure~\ref{distribution}. This allows the network to spread out latent points while still optimizing \eqref{vbound}.  

Additionally, the radius $\Vert z \Vert$ of points drawn from this distribution are near $\tau$ with high probability. Therefore, the norm of an encoded point $\Vert z \Vert$ drawn from $z \sim q_\phi(z|x)$ can be used to create a simple statistic which can be used as an effective test for out of distribution points, see \eqref{S_OOD}. 

At the same time, the tilted Gaussian distribution is obtained as a very simple modification to the ordinary Gaussian. This new prior can easily be implemented in place of the ordinary Gaussian with minimal changes to existing code.


For a tilting parameter $\tau \geq 0$, the exponentially tilted Gaussian distribution, which we denote by $\mathcal{N}_\tau(0,I) \in \mathbb{R}^{d_z}$, is the random variable with density, $\rho_{\tau}:\mathbb{R}^{d_z} \to \mathbb{R}^+$ given by 

\begin{align}
\label{density}
\rho_{\tau}(z) := \frac{e^{\tau \Vert z \Vert}}{Z_{\tau}} \frac{e^{-\frac{1}{2}\Vert z \Vert^2}}{\sqrt{2\pi}^{\frac{d_z}{2}}}.
\end{align}


The additional tilting term $e^{\tau \Vert z \Vert}$ pushes the distribution towards values of greater $\Vert z \Vert$. It is useful to note that by completing the square, the density is proportional to $e^{-\frac{1}{2}\left(||z||-\tau\right)^2}$, meaning that the density is radially symmetric and has a maximum value at $\tau$. Figure~\ref{distribution} shows a illustration of this density.

An important feature of our method is that the encoder distribution (i.e. the distribution assigned in latent space to a single input $x_i$) is \emph{not} exponentially tilted. We take this distribution to be a simple Gaussian of the form $z \sim \mathcal{N}(\mu,I)$ where $\mu = \mu(x)$ is determined by the encoder parameters. Note that for the sake of simplicity and to reduce computation we have fixed the covariance matrix $\Sigma = I$, whereas it is more commonly assumed that $\Sigma = \textrm{diag}(\sigma)$ is allowed to be chosed by the encoder. This simplifying assumption means the tilted prior has less parameters than the standard Gaussian, which results in decreased performance when $\tau \approx 0$. An example of which can be seen in Figure~\ref{cover}.  

For convenience of notation we denote by $f_\tau(\mu)$ the KLD between the encoder and prior distributions, namely

\begin{align}
f_\tau\left(\mu\right) := \infdiv{\mathcal{N}\left(\mu,I\right)}{\mathcal{N}_\tau(0,I)}.
\end{align}

Intuitively, the tilted prior allows the encoder distributions to choose $\mu$ anywhere on the surface of a hyper-sphere with a minimum KLD, rather than located at the single point $\mu = 0$, as would be the case when the prior is an standard Gaussian. Note also that since a Gaussian cannot be perfectly fit to the distribution $\mathcal{N}_\tau(0,I)$, there will always be a non-zero minimum KLD between the encoder and prior distributions, i.e. $\delta(\tau) := \inf_{ \mu \in \mathbb{R}^{d_z} } f_\tau(\mu) > 0$ when $\tau > 0$. This can be interpreted as the minimum average amount of information in nats that each sample contains after being encoded. The decoder is then able to use this information to differentiate distributions in the latent space. This minimum value $\delta(\tau)$ is referred to as the ``committed rate'' in \citet{Razavi}.

\subsection{Normalization Constant and KLD}

The calculation of the distribution's normalization constant $Z_{\tau}$ and $f_\tau(\mu)$ is deffered to Appendices~\ref{appa}-~\ref{appb} and summarized below. The normalizing constant $Z_{\tau}$ can be written as

\begin{align}
Z_{\tau} = &M\left(\frac{d_z}{2},\frac{1}{2},\frac{1}{2}\tau^2\right) \notag\\
&+\tau \sqrt{2} \frac{\Gamma(\frac{d_z+1}{2})}{\Gamma(\frac{d_z}{2})} M\left(\frac{d_z+1}{2},\frac{3}{2},\frac{1}{2}\tau^2\right)
\end{align}

where $M$ stands for the Kummer confluent hypergeometric function $M\left(a,b,z\right) = \sum_{n=0}^{\infty} \frac{a^{\left(n\right)}z^n}{b^{\left(n\right)} n!}$ and $a^{\left(n\right)}$ is the rising factorial $a^{\left(n\right)} = a(a+1)\textrm{. . .}(a+n-1)$.  

The KLD can then be written as 

\begin{equation} \label{eq:f_tau_exact}
f_\tau\left(\mu \right) = \log Z_{\tau} - \tau \sqrt{\frac{\pi}{2}} L_{\frac{1}{2}}^{\frac{d_z}{2}-1} \left(-\frac{\Vert \mu \Vert^2}{2} \right) + \frac{\Vert \mu \Vert^2}{2} 
\end{equation}

where $L$ is the generalized Laguerre polynomial which can be defined as $L_n^{(\alpha)} (x) = \binom{n + \alpha}{n} M\left(-n, \alpha + 1, x\right)$. 

\begin{figure}[ht]
\begin{center}
\centerline{\includegraphics[width=\columnwidth]{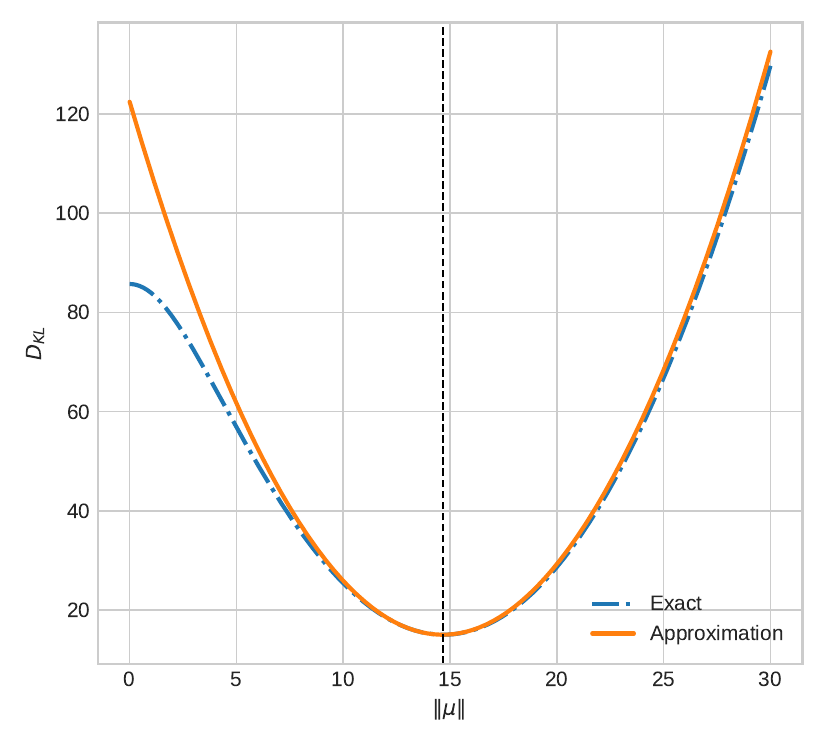}}
\caption{Comparison between the exact and quadratic approximation of the KLD $f_\tau(\mu)$ as in ~\eqref{kldapprox} for $d_z = 10, \tau = 15$}
\label{KLD}
\end{center}
\end{figure}

During training, instead of computing $f_\tau(\mu)$ exactly (which involves doing the infinite summation of the Laguerre polynomial in \eqref{eq:f_tau_exact}), we instead use the following simple quadratic approximation of the KLD which drastically reduces the computation and maintains the  bound in \eqref{vbound}

\begin{align}
\label{kldapprox}
f_\tau\left(\mu\right)\leq\frac{1}{2}\left(\Vert \mu\Vert -\Vert\mu^{\star}(\tau)\Vert\right)^2 + f_\tau\left(\mu^\star(\tau)\right).
\end{align}

where $\mu^\star(\tau) = \textrm{argmin}_{\mu} \textrm{ } f_\tau(\mu)$, which can be arrived at by the proof provided in Appendix~\ref{appc}. Note that $\mu^\star(\tau)$ is not unique and all vectors lying on the hyper-sphere with radius $\Vert\mu^\star\Vert$ are a valid solution. Note also that the term $f_\tau(\mu^\ast(\tau))$ can be omitted during training the VAE since it is a constant. Figure~\ref{KLD} shows an example of the difference between the exact and approximate KLD and Figure~\ref{mu_star} plots $\Vert\mu^\star(\tau)\Vert$ with a few choices of $d_z$.

\newpage

\subsection{Sampling from $\mathcal{N}_\tau(0,I)$}
To generate samples, latent points $z$ are sampled according to the latent distribution $\mathcal{N}_\tau(0,I)$ and then passes them through the decoder. Since the distribution of the latent vector is tilted, this means we must sample for the density $\rho_\tau$ given in \eqref{density}. By spherical symmetry, the \emph{direction} $\frac{z}{\Vert z \Vert}$, is uniformly distributed on the unit sphere and independent of the radius $\Vert z \Vert \in \mathbb{R}$. Therefore the only remaining issue is to sample this radius. 

A simple way to approximately sample $\Vert z \Vert$ is as follows. In Appendix~\ref{appd} we experimentally show that $\Vert z \Vert$ follows the distribution $\mathcal{N}\left(\bar{z}, 1\right)$ where $\bar{z} = \frac{1}{N} \sum_{i=1}^N \Vert z^i \Vert $ and $z\sim q_{\phi}\left(z \vert x\right)$. Given the KLD term in the log-likelihood bound, it is reasonable to expect that the average latent variable length $\bar{z}$ would be approximately equal to $\Vert\mu^\star(\tau)\Vert$, however we consistently observe that $\bar{z} > \Vert\mu^\star(\tau)\Vert$. This implies that longer latent vectors allow the decoder to generate data with lower reconstruction error, resulting in a better optimized log-likelihood bound. 

As a result of these observations, we propose a two step procedure to sample from the model. First generate a vector $u \in\mathbb{R}^{d_z}$ on the unit sphere in $d_z$ dimensions (This can be done efficiently by sampling a standard Gaussian in $\mathbb{R}^{d_z}$, and scaling the resulting vector to length one). Then, scale the length by a sample from the distribution to be length $r\sim\mathcal{N}\left(\bar{z}, 1\right)\in\mathbb{R}$. The resulting sample is:

\begin{figure}[ht]
\begin{center}
\centerline{\includegraphics[width=\columnwidth]{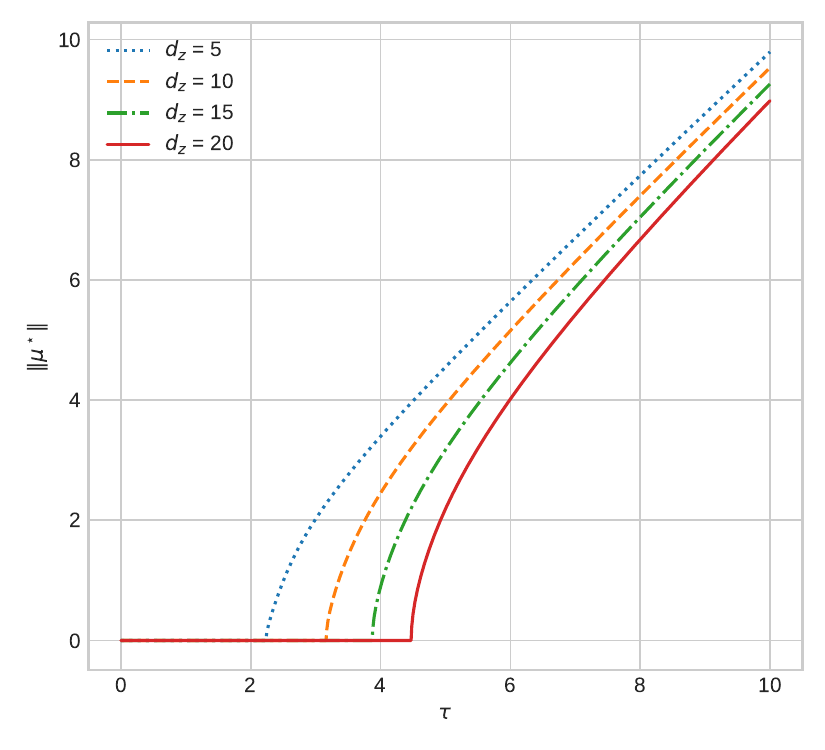}}
\caption{$\Vert\mu^\star(\tau)\Vert$ plotted for a variety of different latent dimensions $d_z$. Note that there is a critical $\tau$ value for which $\mu^\ast(\tau)$ becomes nonzero. }
\label{mu_star}
\end{center}
\end{figure}

\begin{align}
    z_{\text{sample}} = r u \in \mathbb{R}^{d_z}.
\end{align}

\subsection{Out-of-Distribution Detection}
To detect OOD samples, we compute a single score statistic $\mathcal{S}_{\mathrm{OOD}}(x) \in  \mathbb{R}$ for any data point $x \in \mathbb{R}^{d_x}$. Scores less than or equal to some threshold are considered to be in-distribution, and scores greater than the threshold are considered to be out-of-distribution. The score is given by the lower bound on the log-likelihood applied to an encoded data point $x$ in the latent space, $z \sim q_{\phi}(z \vert x) \in \mathbb{R}^{d_z}$ and the distance to the reconstruction $\hat{x} \sim p_\theta(x \vert z) \in \mathbb{R}^{d_x}$,

\begin{align}
\label{S_OOD}
\mathcal{S}_{\mathrm{OOD}}(x) = \Vert x - \hat{x}\Vert + \frac{1}{2}\left(\Vert z\Vert - \Vert\mu^\star(\tau)\Vert\right)^2.
\end{align}

Note that the additive constant in~\eqref{kldapprox} is not included as it would appear as a constant $f_\tau\left(\mu^\star(\tau)\right)$ for all data points.

We choose not to use the importance weighted autoencoder bound (IWAE) \citep{Finke} which is used by a number of current methods \citep{Xiao, Ren}. While the IWAE can achieve a tighter bound on the log-likelihood, it is slower than using the KLD and does not improve the performance of detecting OOD samples with the tilted prior. We use the $l_2$ reconstruction loss to calculate $\mathbb{E}_{q_{\phi}(z \vert x)} \textrm{log }\left[p_{\theta}(x \vert z)\right]$. 

\section{Related Work}
\subsection{Alternative Priors for Variational Autoencoders}

Arguably the most commonly used prior distribution for the VAE is the standard Gaussian, which takes the form $\mathcal{N}(0,\textrm{diag}(\sigma) )$ where $\sigma = \left[\sigma_0 \textrm{. . . } \sigma_{n-1}\right]^\top$. While this formulations has seen great popularity, there are a number of drawbacks such as the posterior collapse problem and blurry image reconstructions due to the Gaussian distribution being uni-modal. In this section we discuss a variety of different prior distributions that have been proposed as an alternative to the standard Gaussian prior.

A popular method is to use a hierarchy of variables for the prior distribution. For example the Ladder VAE constructs a hierarchy of the form $p_{\theta}=p_{\theta}\left(z_L\right)\prod_{i=1}^{L-1} p_{\theta}\left(z_i\vert z_{i+1}\right)$ where the hierarchy has $L$ layers of latent variable. Some other example of this type of model include work done by \citet{Sonderby}, \citet{Maaloe1}, and \citet{Maaloe2} which all use different ways to construct the latent variable hierarchy. 

Another type of prior can be constructed by using flow-based methods in the latent space of a VAE. This was originally proposed by \citet{Rezende} with planar and radial flows as potential candidate flow methods. Many methods have been proposed that build on this idea, for example \citet{Tomczak} which creates flows with the Householder transformation while \citet{Kingma1} uses an autoregressive model.

The last type of method we review uses a ``committed rate" which is introduced in \citep{Razavi} and refers to a lower bound on the KLD of the form $D_{\mathrm{KL}}\geq\delta$ where $\delta$ is a flexible parameter. This has the advantage of encoding more information in the latent space and prevents posterior collapse \textit{a priori}. They use a correlated sequential model to achieve their flexible committed rate. A different line of work uses the vMF distribution $\textrm{vMF}(\mu,\kappa)$ with a fixed $\kappa$ as a posterior and $\textrm{vMF}(\cdot,0)$ as the prior. A drawback to this method is that is does not allow higher KLD for different points, something which has been shown to be useful for detecting outliers \citep{vandenoord1}.

\begin{figure*}[ht]
\begin{center}
\centerline{\includegraphics[scale=0.66]{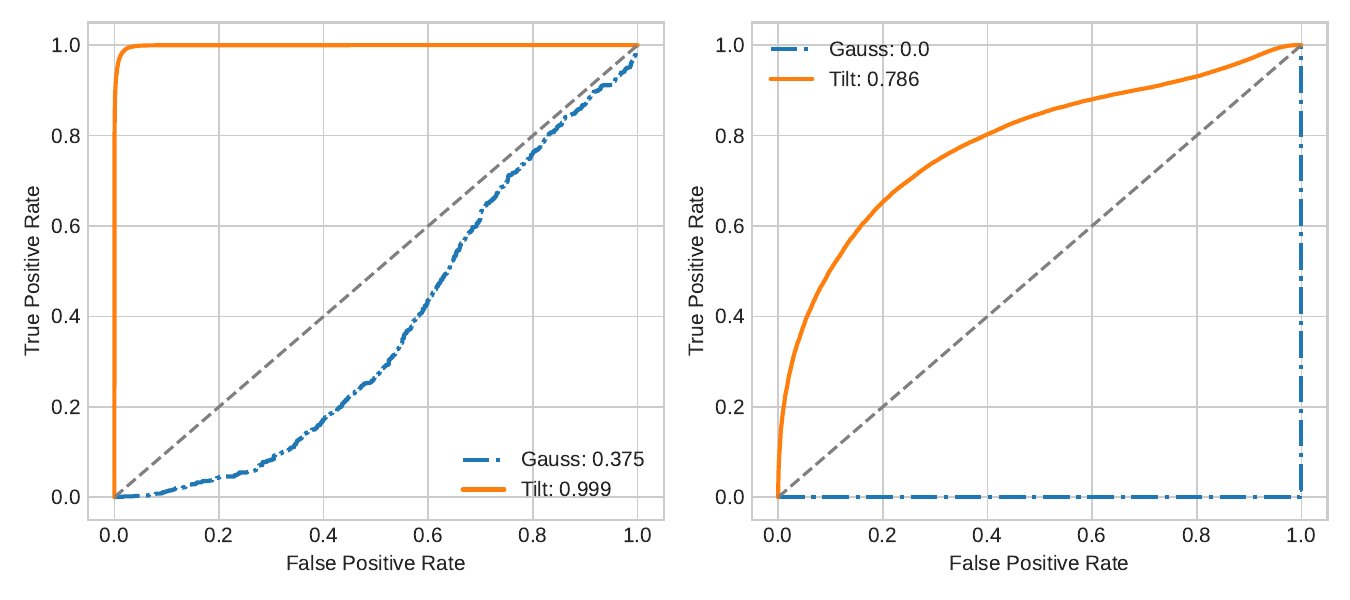}}
\caption{A comparison of the ROC of models trained with a standard Gaussian prior and an tilted prior. The left plot is data from models trained on Fashion-MNIST and the right plot is data from models trained on CIFAR-10. Both cases are tested for out of distribution detection performance on the MNIST dataset. The plot is obtained by varying the threshold score to classify as in or out of distribution. The numbers in the legend are the AUCROC score.}
\label{ROC}
\end{center}
\end{figure*}

\subsection{Unsupervised Out-of-Distribution Detection Methods}
A variety of methods have been proposed to improve the performance of unsupervised methods on OoD detection. Initially, it was proposed that a one-sided log-likelihood test would be a sufficient test statistic \citep{SChoi}. Since the discovery that this methodology appears to fail on many different types of generative models, many alternative scores have been proposed. Likelihood Ratios \citep{Ren} proposes to use the ratio between two different types of networks, one that captures the semantic content of data, and the other which captures background information. Likelihood Regret \citep{Xiao} uses a similar principle, but uses the ratio between the optimal network for the training data and the optimal network for an individual sample. ROSE \citep{HChoi} uses a method to compute how much a sample would update a model's parameters. One drawback of these methods is that the processes are significantly slower than the original VAE forward pass. 

Input complexity \citep{Serra} shows that OOD performance can be improved by using an estimate of the Kolmogorov complexity alongside the one-sided log-likelihood test. In the case of images the authors propose to use popular image compression algorithms like JPG2000 as the complexity estimate, which achieves excellent performance on  both normalizing flows and VAEs. Density of States Estimation \citep{Morningstar} uses a approach based on statistical physics and directly measures the typicality of different model statistics to classify OOD samples. 

Some new scores use batch properties to generate better results. \citet{Nalisnick2} uses a typicality test which requires at least a pair of images per batch to function and \citet{Song} uses batch normalization statistics. Although both works have impressive results, difficulties with using these methods occur due to the batch properties. In the case of \citet{Nalisnick2} training data and OOD samples must be fully separated in batches, which is impractical for many applications. For the method proposed by \citet{Song} training and OOD data can be mixed in batches, however the detection performance is tied to the ratio of in distribution vs OOD data in each batch.

A different class of methods uses OOD samples in the training process along with a procedure to encourage a network to learn to differentiate between OOD samples and the training data. Examples of this type of training include \citet{Lee} who uses a GAN generator to create OOD samples and \citet{Hendrycks} who uses a completely disjoint dataset to train with, for example CIFAR-10 trained with CIFAR-100 as OOD. Some papers use perturbations of input images to create data that is treated as OOD. \citet{SChoi} uses blurred images as adversarial examples, and 
and \citet{Ran} adds Gaussian noise to images and employs a noise contrastive prior for the VAE architecture.

Finally \citet{Havtorn} and \citet{Maaloe2} make use of a hierarchical latent variable prior for the VAE. They show that the hierarchical prior improves the VAE's ability to perform OOD detection. These works point to the possibility of performing unsupervised OOD detection with VAE's using just the log-likelihood as a test statistic, and without any additional perturbed training data.

\section{Results}

\begin{table*}[ht]
\caption{AUROC comparison between OOD detection methods with Fashion-MNIST and CIFAR-10 as training distributions. A larger number is better. }
\label{AUROC-table}
\vskip 0.15in
\begin{center}
\begin{small}
\begin{sc}
\begin{tabular}{lccccc|cc}
\toprule
Dataset & LL & IC (png) & IC (JPEG2000) & Ratio & Regret & Tilt \\
\midrule
\underline{Fashion-MNIST}    \\
MNIST           & 0.375 & 0.993 & 0.351 & 0.965 & \bf{0.999} & \bf{0.999} \\
CIFAR-10        & \bf{1.0} & 0.97 & \bf{1.0} & 0.914 & 0.986 & 0.997 \\
SVHN            & \bf{1.0} & 0.999 & \bf{1.0} & 0.761 & 0.989 & 0.98 \\
KMNIST          & 0.765 & 0.863 & 0.769 & 0.96 & 0.998 & \bf{0.999} \\
Noise           & \bf{1.0} & 0.324 & \bf{1.0} & \bf{1.0} & 0.998 & \bf{1.0} \\
Constant        & 0.975 & \bf{1.0} & 0.984 & 0.98 & 0.999 & 0.798 \\
\hline
\\
\underline{CIFAR-10}       \\
MNIST           & 0.0 & 0.976 & 0.0 & 0.032 & \bf{0.986} & 0.797 & \\
Fashion-MNIST   & 0.032 & \bf{0.987} & 0.035 & 0.335 & 0.976 & 0.688 & \\
SVHN            & 0.209 & \bf{0.938} & 0.215 & 0.732 & 0.912 & 0.143 & \\
LSUN            & 0.833 & 0.348 & 0.833 & 0.508 & 0.606 & \bf{0.933} & \\
CelebA          & 0.676 & 0.310 & 0.679 & 0.404 & 0.738 & \bf{0.877} & \\
Noise           & \bf{1.0} & 0.042 & \bf{1.0} & 0.851 & 0.994 & \bf{1.0} & \\
Constant        & 0.015 & \bf{1.0} & 0.269 & 0.902 & 0.974 & 0.0 & \\

\bottomrule
\end{tabular}
\end{sc}
\end{small}
\end{center}
\vskip -0.1in
\end{table*}

First, we conduct a basic experiment demonstrating the effect of using the tilted prior compared to the standard Gaussian prior. Details regarding all experimental settings can be found in Appendix~\ref{appe}\footnote{Code is available at \url{https://github.com/anonconfsubaccount/tilted_prior}}. To give a quantitative measure of OOD classification, we examine the Receiver Operating Characteristics (ROC) curve which looks at how the false positive and true positive rates are related. We use the Area Under the Curve-Receiver Operating Characteristics (AUCROC) as a metric \citep{fawcett}. Another metric, the False Positive Rate at 80\% True Positive Rate (FPR80) is also displayed in Appendix~\ref{appf}.

When a VAE with a standard Gaussian prior is trained on either the Fashion-MNIST or CIFAR-10 dataset, we observe that the OOD MNIST dataset is scored with a higher log-likelihood than the training datasets in both cases. The resulting AUCROC score is less than 0.5 for both models, which corresponds to poorer performance than a random classifier. By using our proposed tilted prior for this task, we observe that the VAEs give a more accurate score to the OOD MNIST dataset. The ROC curves are plotted for both experiments in Figure~\ref{ROC}. The tilted prior improves performance for the AUCROC metric on the Fashion-MNIST test from 0.375 to 0.999 and from 0 to 0.73 on the CIFAR-10 test. Importantly, this type of failure when using the standard Guassian prior is not limited to these datasets, but is a pervasive issue that occurs in a large number of OOD datasets.

\subsection{Comparison to other Methods on Out-of-Distribution Detection}

To further validate the performance of our proposed prior, we compare the tilted prior to a variety of recent methods that achieve top performance in OOD detection for VAE's. The methods we compare against are a VAE with a standard Gaussian prior (\textbf{LL}), Input Complexity (\textbf{IC (png)}), (\textbf{IC (JPEG2000)}) \citep{Serra}, Likelihood Ratios (\textbf{Ratio}) \citep{Ren}, and
Likelihood Regret (\textbf{Regret}) \citep{Xiao}. The results from this analysis are summarized on Table~\ref{AUROC-table}. 

Overall, we observe that the tilted prior performs well in comparison to the other OOD detection methods. In 6 of the test cases our method achieve top performance compared to the next best method Input Complexity, which achieves top performance in 4 of the test cases. A failure case of our method occurs when the complexity of OOD samples are much less than that of the training data. We follow \citep{Serra} and define complexity as an upper bound of the Kolmogorov Complexity, which can be calculated with a lossless compression algorithm. Examples of the low-complexity OOD failure case occur for the CIFAR-10 vs Constant and the CIFAR-10 vs SVHN tests.  Our best performance is achieved when the training and OOD complexity are roughly similar, as is the case for CIFAR-10 vs LSUN and Fashion-MNIST vs KMNIST tests.

\subsection{Speed Comparison}
In this section we compare the runtime performance of the OOD detection models in the previous section. Following \citet{Xiao} we use IWAE with 200 samples to derive a lower bound on the log-likelihood IWAE samples for Likelihood Regret, Likelihood Ratio, and Input Complexity which are computed in a single batch. The tilted prior computes the KLD and reconstruction error in one sample, meaning that it is $\mathcal{O}(k)$ faster than methods which use IWAE, where $k$ is the number of IWAE samples. To include a more fair test, we also show the images per second for our method under the scenario where $k$ latent vectors are sampled from the encoder which are used to obtain a more accurate estimate of $\mathbb{E}_{q_{\phi}(z \vert x)}\left[\textrm{log }p_{\theta}(x \vert z)\right]$. We label this test \textbf{Tilt (Batch)}. 

We observe that the tilted prior has significantly faster runtime than other methods that are used to detect OOD images. In the instance where the tilted prior is exploiting sample efficiency speed-ups, we see runtime improvements of up to 4 orders of magnitude. This makes our method appealing for applications where power and timing requirements are important.

\begin{table}[ht]
\caption{Runtime comparison between methods on the Fashion-MNIST dataset. A larger number is better. The test was performed using a NVIDIA RTX 3090 and an AMD Ryzen 7 3800X}
\label{speed-table}
\vskip 0.15in
\begin{center}
\begin{small}
\begin{sc}
\begin{tabular}{lcccr}
\toprule
Method & Images per Second \\
\midrule
Tilt            & $\mathbf{3.22\times 10^4}$ \\
Tilt (batch)    & $1.61\times 10^2$ \\

\hline
\\
IC (PNG)        & $1.02\times 10^2$ \\
IC (JP2)        & $1.00\times 10^2$ \\
Ratio           & $4.26\times 10^1$ \\
Regret          & $2.51\times 10^0$ \\

\bottomrule
\end{tabular}
\end{sc}
\end{small}
\end{center}
\vskip -0.1in
\end{table}

\subsection {Sample Quality and Log-Likelihood}
Here we compare the sample quality and log-likelihood of the tilted prior to the  standard Gaussian prior. We observe that the standard Gaussian prior consistently achieves a larger log-likelihood than our method. This is due to the large committed rate $f_\tau(\mu) \geq \delta(\tau)$ which the tilted prior has when $\tau\gg 0$. Furthermore, given that the tilted prior lacks a learnable factorized covariance matrix $\Sigma=\textrm{diag}(\sigma)$, the standard Gaussian prior is able to out-perform the tilted prior on the log-likelihood lower bound when $\tau\approx 0$.

Despite having low scores on the log-likelihood bound, the tilted prior still functions well as a generative model. On the task of sample reconstruction we observe that the tilted prior has significantly lower error than the standard Gaussian prior. While this is also possible with an autoencoder, our method provides high quality samples from the latent space, something that is difficult for auto-encoders to do given the lack of structure in the latent space. Figures~\ref{MNIST-recon}-\ref{MNIST-sample} contain examples of reconstructions and samples respectively, on VAEs trained with the Fashion-MNIST dataset.

We observe that samples from the tilted prior appear to be more crisp than samples from the standard Gaussian prior, which are known to be ``fuzzy" due to the uni-modal prior. Furthermore, we observe that some data classes are completely absent from the standard Gaussian VAEs latent space which exist when using the tilted prior. Example of this can be seen in Figure~\ref{MNIST-recon} where the standard Gaussian VAE is unable to reconstruct the high-heel shoe type or handbag, whereas the tilted prior can.

\section{Discussion}
In this work we have introduced a novel prior distribution for the VAE. By testing on the established benchmark we determined that the VAE with a tilted prior distribution can achieve state-of-the-art results on OOD detection using the natural log-likelihood test statistic. Furthermore, we show that the prior has a simple implementation and fast runtime. When sampling from the model, we find that the produced images are of high quality compared to those produced from the standard Gaussian prior. Our work supports the hypothesis that it is possible for VAEs to reliably perform unsupervised OOD detection without introducing new scores. We hope that this work can lead to further developments to improve upon the OOD detection task for VAEs.
\bibliography{tilted_prior}

\begin{thebibliography}{29}
\providecommand{\natexlab}[1]{#1}
\providecommand{\url}[1]{\texttt{#1}}
\expandafter\ifx\csname urlstyle\endcsname\relax
  \providecommand{\doi}[1]{doi: #1}\else
  \providecommand{\doi}{doi: \begingroup \urlstyle{rm}\Url}\fi

\bibitem[Alemi et~al.(2018)Alemi, Fischer, and Dillon]{Alemi}
Alemi, A.~A., Fischer, I., and Dillon, J.~V.
\newblock Uncertainty in the variational information bottleneck.
\newblock In \emph{arXiv preprint arXiv:1807.00906}, 2018.

\bibitem[Amodei et~al.(2016)Amodei, Olah, Steinhardt, Christiano, Schulman, and
  Mané]{Amodei}
Amodei, D., Olah, C., Steinhardt, J., Christiano, P., Schulman, J., and Mané,
  D.
\newblock Concrete problems in ai safety.
\newblock In \emph{arXiv preprint arXiv:1606.06565}, 2016.

\bibitem[Choi et~al.(2019)Choi, Jang, and Alemi]{HChoi}
Choi, H., Jang, E., and Alemi, A.~A.
\newblock Wic, but why? generative ensembles for robust anomaly detection.
\newblock In \emph{arXiv preprint arXiv:1810.01392}, 2019.

\bibitem[Choi \& Chung(2020)Choi and Chung]{SChoi}
Choi, S. and Chung, S.-Y.
\newblock Novelty detection via blurring.
\newblock In \emph{International Conference on Learning Representations}, 2020.

\bibitem[Fawcett(2006)]{fawcett}
Fawcett, T.
\newblock Introduction to roc analysis.
\newblock 2006.

\bibitem[Finke \& Thiery(2019)Finke and Thiery]{Finke}
Finke, A. and Thiery, A.~H.
\newblock On importance-weighted autoencoders.
\newblock In \emph{arXiv preprint arXiv:1907.10477}, 2019.

\bibitem[Havtorn et~al.(2021)Havtorn, Frellsen, Hauberg, and
  Maal{\o}e]{Havtorn}
Havtorn, J. D.~D., Frellsen, J., Hauberg, S., and Maal{\o}e, L.
\newblock Hierarchical vaes know what they don’t know.
\newblock In \emph{Proceedings of the 38th International Conference on Machine
  Learning}, pp.\  4117--4128, 2021.

\bibitem[Hendrycks et~al.(2019)Hendrycks, Mazeika, and Dietterich]{Hendrycks}
Hendrycks, D., Mazeika, M., and Dietterich, T.
\newblock Deep anomaly detection with outlier exposure.
\newblock In \emph{International Conference on Learning Representations}, 2019.

\bibitem[Hoffman \& Johnson(2016)Hoffman and Johnson]{Hoffman}
Hoffman, M. and Johnson, M.
\newblock Elbo surgery: yet another way to carve up the variational evidence
  lower bound.
\newblock In \emph{Advances in Approximate Bayesian Inference, NIPS Workshop},
  2016.

\bibitem[Kingma \& Dhariwal(2018)Kingma and Dhariwal]{Kingma3}
Kingma, D.~P. and Dhariwal, P.
\newblock Glow: Generative flow with invertible 1x1 convolutions.
\newblock In \emph{Advances in Neural Information Processing Systems}, 2018.

\bibitem[Kingma \& Welling(2014)Kingma and Welling]{Kingma2}
Kingma, D.~P. and Welling, M.
\newblock Auto-encoding variational bayes.
\newblock In \emph{arXiv preprint arXiv:1312.6114}, 2014.

\bibitem[Kingma et~al.(2016)Kingma, Salimans, Jozefowicz, Chen, Sutskever, and
  Welling]{Kingma1}
Kingma, D.~P., Salimans, T., Jozefowicz, R., Chen, X., Sutskever, I., and
  Welling, M.
\newblock Improved variational inference with inverse autoregressive flow.
\newblock In \emph{Advances in Neural Information Processing Systems}, 2016.

\bibitem[Lee et~al.(2018)Lee, Lee, Lee, and Shin]{Lee}
Lee, K., Lee, H., Lee, K., and Shin, J.
\newblock Training confidence-calibrated classifiers for detecting
  out-of-distribution samples.
\newblock In \emph{International Conference on Learning Representations}, 2018.

\bibitem[Maaløe et~al.(2016)Maaløe, Sønderby, Sønderby, and
  Winther]{Maaloe1}
Maaløe, L., Sønderby, C.~K., Sønderby, S.~K., and Winther, O.
\newblock Auxiliary deep generative models.
\newblock In \emph{Proceedings of The 33rd International Conference on Machine
  Learning}, pp.\  1445--1453, 2016.

\bibitem[Maaløe et~al.(2019)Maaløe, Fraccaro, Li\'{e}vin, and
  Winther]{Maaloe2}
Maaløe, L., Fraccaro, M., Li\'{e}vin, V., and Winther, O.
\newblock Biva: A very deep hierarchy of latent variables for generative
  modeling.
\newblock In Wallach, H., Larochelle, H., Beygelzimer, A., d\textquotesingle
  Alch\'{e}-Buc, F., Fox, E., and Garnett, R. (eds.), \emph{Advances in Neural
  Information Processing Systems}, 2019.

\bibitem[Morningstar et~al.(2021)Morningstar, Ham, Gallagher, Lakshminarayanan,
  Alemi, and Dillon]{Morningstar}
Morningstar, W., Ham, C., Gallagher, A., Lakshminarayanan, B., Alemi, A., and
  Dillon, J.
\newblock Density of states estimation for out of distribution detection.
\newblock In \emph{Proceedings of The 24th International Conference on
  Artificial Intelligence and Statistics}, Proceedings of Machine Learning
  Research, pp.\  3232--3240, 2021.

\bibitem[Nalisnick et~al.(2019{\natexlab{a}})Nalisnick, Matsukawa, Teh, Gorur,
  and Lakshminarayanan"]{Nalisnick1}
Nalisnick, E., Matsukawa, A., Teh, Y.~W., Gorur, D., and Lakshminarayanan", B.
\newblock Do deep generative models know what they don't know?
\newblock In \emph{International Conference on Learning Representations},
  2019{\natexlab{a}}.

\bibitem[Nalisnick et~al.(2019{\natexlab{b}})Nalisnick, Matsukawa, Teh, and
  Lakshminarayanan"]{Nalisnick2}
Nalisnick, E., Matsukawa, A., Teh, Y.~W., and Lakshminarayanan", B.
\newblock Detecting out-of-distribution inputs to deep generative models using
  typicality.
\newblock In \emph{arXiv preprint arXiv:1906.02994}, 2019{\natexlab{b}}.

\bibitem[Ran et~al.(2021)Ran, Xu, Mei, Xu, and Liu]{Ran}
Ran, X., Xu, M., Mei, L., Xu, Q., and Liu, Q.
\newblock Detecting out-of-distribution samples via variational auto-encoder
  with reliable uncertainty estimation.
\newblock In \emph{arXiv preprint arXiv:2007.08128}, 2021.

\bibitem[Razavi et~al.(2019)Razavi, van~den Oord, Poole, and Vinyals]{Razavi}
Razavi, A., van~den Oord, A., Poole, B., and Vinyals, O.
\newblock Preventing posterior collapse with delta-vaes.
\newblock In \emph{International Conference on Learning Representations}, 2019.

\bibitem[Ren et~al.(2019)Ren, Liu, Fertig, Snoek, Poplin, Depristo, Dillon, and
  Lakshminarayanan]{Ren}
Ren, J., Liu, P.~J., Fertig, E., Snoek, J., Poplin, R., Depristo, M., Dillon,
  J., and Lakshminarayanan, B.
\newblock Likelihood ratios for out-of-distribution detection.
\newblock In \emph{Advances in Neural Information Processing Systems}, 2019.

\bibitem[Rezende \& Mohamed(2015)Rezende and Mohamed]{Rezende}
Rezende, D. and Mohamed, S.
\newblock Variational inference with normalizing flows.
\newblock In \emph{Proceedings of the 32nd International Conference on Machine
  Learning}, pp.\  1530--1538, 2015.

\bibitem[Serrà et~al.(2020)Serrà, Álvarez, Gómez, Slizovskaia, Núñez, and
  Luque]{Serra}
Serrà, J., Álvarez, D., Gómez, V., Slizovskaia, O., Núñez, J.~F., and
  Luque, J.
\newblock Input complexity and out-of-distribution detection with
  likelihood-based generative models.
\newblock In \emph{International Conference on Learning Representations}, 2020.

\bibitem[Song et~al.(2019)Song, Song, and Ermon]{Song}
Song, J., Song, Y., and Ermon, S.
\newblock Unsupervised out-of-distribution detection with batch normalization.
\newblock In \emph{arXiv preprint arXiv:1910.09115}, 2019.

\bibitem[Sønderby et~al.(2016)Sønderby, Raiko, Maal\o~e, S\o~nderby, and
  Winther]{Sonderby}
Sønderby, C.~K., Raiko, T., Maal\o~e, L., S\o~nderby, S. r.~K., and Winther,
  O.
\newblock Ladder variational autoencoders.
\newblock In Lee, D., Sugiyama, M., Luxburg, U., Guyon, I., and Garnett, R.
  (eds.), \emph{Advances in Neural Information Processing Systems}, 2016.

\bibitem[Tomczak \& Welling(2017)Tomczak and Welling]{Tomczak}
Tomczak, J.~M. and Welling, M.
\newblock Improving variational auto-encoders using householder flow.
\newblock In \emph{arXiv preprint arXiv:1611.09630}, 2017.

\bibitem[van~den Oord et~al.(2016)van~den Oord, Kalchbrenner, Espeholt,
  kavukcuoglu, Vinyals, and Graves]{vandenoord2}
van~den Oord, A., Kalchbrenner, N., Espeholt, L., kavukcuoglu, k., Vinyals, O.,
  and Graves, A.
\newblock Conditional image generation with pixelcnn decoders.
\newblock In \emph{Advances in Neural Information Processing Systems}, 2016.

\bibitem[van~den Oord et~al.(2017)van~den Oord, Vinyals, and
  kavukcuoglu]{vandenoord1}
van~den Oord, A., Vinyals, O., and kavukcuoglu, k.
\newblock Neural discrete representation learning.
\newblock In \emph{Advances in Neural Information Processing Systems}, 2017.

\bibitem[Xiao et~al.(2020)Xiao, Yan, and Amit]{Xiao}
Xiao, Z., Yan, Q., and Amit, Y.
\newblock Likelihood regret: An out-of-distribution detection score for
  variational auto-encoder.
\newblock In \emph{Advances in Neural Information Processing Systems}, pp.\
  20685--20696, 2020.

\end{thebibliography}
\bibliographystyle{icml2022}

\newpage
\appendix
\onecolumn
\appendix
\section{Derivation of the Normalization Constant for the Tilted Gaussian}
\label{appa}
\begin{align*}
\rho_{\tau}(z) &= \frac{e^{\tau \Vert z \Vert}}{Z_{\tau}} \frac{e^{-\frac{1}{2}\Vert z \Vert^2}}{\sqrt{2\pi}^{d/2}} dz\\
Z_{\tau} & = \mathbb{E}_{z\sim\mathcal{N}(0,I)}\left[e^{\tau\Vert z\Vert}\right]=\int_{\mathbb{R}^{d}}e^{\tau \Vert z\Vert}\frac{e^{-\frac{1}{2}\Vert z\Vert^2}}{\sqrt{2\pi}^{d/2}}\mathrm{d}z\\
& =\mathbb{E}_{x\sim\chi(d)}\left[e^{\tau x}\right]=\int_{0}^{\infty}e^{\tau
x}\frac{x^{d-1}e^{-\frac{1}{2} x^2}}{2^{\frac{1}{2} d-1}\Gamma\left(\frac{d}{2}\right)}\mathrm{d}x,\\
&\text{ where }\chi\stackrel{d}{=}\Vert z \Vert\\
& =\sum_{n=0}^{\infty}\frac{\tau^{n}\mathbb{E}\left[\chi^{n}\right]}{n!}\\
& =\sum_{n\text{ even}}^{\infty}\frac{\tau^{n}d\left(d+2\right)\ldots\left(d+n-2\right)}{n!}\\
& +\sum_{n\text{ odd}}^{\infty}\frac{\tau^{n}\mu_{1}\left(d+1\right)\left(d+3\right)\ldots\left(d+n-2\right)}{n!}\\
& \text{ where }\mu_1=\mathbb{E}\left[\chi\right]=\sqrt{2}\frac{\Gamma\left(\frac{d+1}{2}\right)}{\Gamma\left(\frac{d}{2}\right)}\\
& =M\left(\frac{d}{2},\frac{1}{2},\frac{1}{2}\tau^{2}\right)\\
&+\tau\sqrt{2}\frac{\Gamma\left(\frac{d+1}{2}\right)}{\Gamma\left(\frac{d}{2}\right)}M\left(\frac{d+1}{2},\frac{3}{2},\frac{1}{2}\tau^{2}\right) \\
\end{align*}

where $M\left(a,b,z\right)=\sum_{n=0}^{\infty}\frac{a^{\left(n\right)}}{b^{\left(n\right)}}\frac{z^n}{n!}$ is the Kummer confluent hypergeometric function and $a^{\left(n\right)} = a(a+1)\ldots(a+n-1)$ is the rising factorial.

\section{Derivation of the KLD}
\label{appb}
\begin{align*}
f_{\tau}(\mu) &= \infdiv{\mathcal{N}(\mu,I)}{\mathcal{N}_\tau(0,I)}\\
& =\mathbb{E}_{z\sim\mathcal{N}(\mu,I)}\left[\ln\left(\frac{\rho_{\mathcal{N}(\mu,I)}(z)}{\rho_{\mathcal{N}_\tau(0,I)}(z)}\right)\right]\\
& =\mathbb{E}_{z\sim\mathcal{N}(\mu,I)}\left[\ln\left(\frac{e^{-\frac{1}{2}{\Vert z -\mu\Vert}^{2}}}{\frac{1}{Z_{\tau}}e^{\tau\Vert z\Vert}e^{-\frac{1}{2}\Vert z\Vert^{2}}}\right)\right]\\
& =\mathbb{E}_{z\sim\mathcal{N}(\mu,I)}\\
& \left[\ln\left(Z_{\tau}e^{-\tau\Vert z\Vert}e^{-\frac{1}{2}\Vert z\Vert^2+\left\langle z,\mu\right\rangle -\frac{1}{2}\Vert\mu\Vert^2+\frac{1}{2}\Vert z\Vert^2}\right)\right]\\
& =\ln(Z_{\tau})-\tau\mathbb{E}_{z\sim\mathcal{N}(\mu,I)}\left[\Vert z\Vert\right]+\frac{1}{2}\Vert\mu\Vert^2\\
& =\ln(Z_{\tau})-\tau L_{\frac{1}{2}}^{(d/2-1)}\left(-\frac{\Vert\mu\Vert^2}{2}\right) +\frac{1}{2}\Vert\mu\Vert^2\text,\\
\end{align*}
 
where $\mathbb{E}_{z\sim\mathcal{N}(\mu,I)}\left[\Vert z\Vert\right] = \mathbb{E}[\chi_{\mu}] = L_{\frac{1}{2}}^{(d/2-1)}\left(-\frac{\Vert\mu\Vert^2}{2}\right)$ and $L_{n}^{(\alpha)}$ is the generalized Laguerre polynomial.

\section{KLD Approximation}
\label{appc}
Let $\tau\in\mathbb{R}$ be a parameter and define $f_{\tau}:\mathbb{R}^{d}\to\mathbb{R}$
by
\begin{align*}
f_{\tau}(\mu) & = \infdiv{\mathcal{N}(\mu,I)}{\mathcal{N}(0,\tau)}\\
& =\ln(Z_{\tau})-\tau\mathbb{E}_{z\sim\mathcal{N}(\mu,I)}\left[\norm z\right]+\frac{1}{2}\norm{\mu}^{2}\\
\text{(use }\mathcal{N}(\mu,I)=\mu+\mathcal{N}(0,I)) & =\ln(Z_{\tau})-\tau\mathbb{E}_{z\sim\mathcal{N}(0,I)}\left[\norm{z+\mu}\right]+\frac{1}{2}\norm{\mu}^{2}
\end{align*}

Note that $f_{\tau}$ depends only on the magnitude of $\mu$, not
the direction. With $x=\norm{\mu},$ we can therefore define $g_{\tau}:\mathbb{R}\to\mathbb{R}$
by
\[
g_{\tau}(x):=\ln(Z_{\tau})-\tau\mathbb{E}_{z\sim\mathcal{N}(0,I)}\left[\norm{z+x\vec{e}_{1}}\right]+\frac{1}{2} x^{2}
\]
where $\vec{e}_{1}$ is the unit vector $\vec{e}_{1}=\left(1,0,0,\ldots,0\right)^{T}$.

\begin{proposition}
The function $g$ satisfies
\[
g_{\tau}^{\prime\prime}(x)=1-\tau\mathbb{E}_{z\sim\mathcal{N}(0,I)}\left[\frac{z_{2}^{2}+\ldots+z_{d}^{2}}{\norm{z+x\vec{e}_{1}}^{3}}\right]
\]

and in particular for $\tau>0$, 
\[
g_{\tau}^{\prime\prime}(x)\leq1
\]
\end{proposition}

\begin{proof}
The proof goes by taking the derivative of $g_{\tau}(x):=\ln(Z_{\tau})-\tau\mathbb{E}_{z\sim\mathcal{N}(0,I)}\left[\norm{z+x\hat{u}}\right]+\frac{1}{2} x^{2}$
and observing that
\[
\frac{\textrm{d}^{2}}{\textrm{d} x^{2}}\norm{z+x\vec{e}_{1}}=\frac{z_{2}^{2}+\ldots+z_{d}^{2}}{\norm{z+x\vec{e}_{1}}^{3}},
\]

which is exactly the claimed Type II Beta distribution. To check the
derivative, we use the fact that $\frac{\textrm{d}}{\textrm{d} x}\norm{z+x\vec{e}_{1}}^{2}=2(z_{1}+x)$
and so:
\begin{align*}
\frac{\textrm{d}}{\textrm{d} x}\norm{z+x\vec{e}_{1}} & =\frac{\textrm{d}}{\textrm{d} x}\left(\norm{z+x\vec{e}_{1}}^{2}\right)^{\frac{1}{2}}\\
 & =\frac{1}{2}\left(\norm{z+x\vec{e}_{1}}^{2}\right)^{-\frac{1}{2}}2\left(z_{1}+x\right)\\
 & =\left(\norm{z+x\vec{e}_{1}}^{2}\right)^{-\frac{1}{2}}\left(z_{1}+x\right)
\end{align*}
 and so taking another derivative gives:
\begin{align*}
\frac{\textrm{d}^{2}}{\textrm{d} x^{2}}\norm{z+x\vec{e}_{1}} & =-\frac{1}{2}\left(\norm{z+x\vec{e}_{1}}^{2}\right)^{-\frac{3}{2}}2(z_{1}+x)\left(z_{1}+x\right)\\
 & \ +\left(\norm{z+x\vec{e}_{1}}^{2}\right)^{-\frac{1}{2}}\cdot1\\
 & =\left(\norm{z+x\vec{e}_{1}}^{2}\right)^{-\frac{3}{2}}\left(-\left(z_{1}+x\right)^{2}+\norm{z+x\vec{e}_{1}}^{2}\right)\\
 & =\frac{z_{2}^{2}+\ldots+z_{d}^{2}}{\norm{z+x\vec{e}_{1}}^{3}}
\end{align*}
\end{proof}
\begin{remark}
Note that the random variable appearing in the formula for $g_{\tau}$
is closely related to the non-central Beta distribution 
\[
X\left(\frac{m}{2},\frac{n}{2},\lambda\right)\stackrel{d}{=}\frac{\sum_{i=n+1}^{n+m}z_{i}^{2}}{\left(z_{1}+\lambda\right)^{2}+\sum_{i=2}^{m}z_{i}^{2}+\sum_{i=n+1}^{n+m}z_{j}^{2}}\in(0,1)
\]
\end{remark}

\begin{corollary}
Let $\mu^{\ast}(\tau)$ be the minimum of $f_{\tau}$ so that $f_{\tau}\left(\mu^{\ast}(\tau)\right)\leq f_{\tau}\left(\mu\right)$
for all $\mu\in\mathbb{R}^{d}$ and $g_{\tau}(\norm{\mu^{\ast}(\tau)})\leq g_{\tau}(x$)
for all $x\in\mathbb{R}$. Then:
\[
g_{\tau}(x)\leq\frac{1}{2}\left(x-\norm{\mu^{\ast}(\tau)}\right)^{2}+g_{\tau}\left(\norm{\mu^{\ast}(\tau)}\right)
\]

or equivalently:
\[
f_{\tau}(\mu)\leq\frac{1}{2}\left(\norm{\mu}-\norm{\mu^{\ast}(\tau)}\right)^{2}+g_{\tau}\left(\norm{\mu^{\ast}(\tau)}\right)
\]
\end{corollary}

\begin{proof}
Since $g_{\tau}^{\prime\prime}(x)=1-\tau\mathbb{E}\left[X\left(\frac{d}{2},\frac{1}{2};x\right)\right]$
where $X$ is a non-central Beta random variable satisfying $X\in\left[0,1\right]$,
it is clear that $g_{\tau}^{\prime\prime}(x)\leq1-\tau\cdot0\leq1$
. Since $\norm{\mu^{\ast}(\tau)}$ is the minimum and $g$ is differentiable,
we also have $g_{\tau}^{\prime}(\norm{\mu^{\ast}(\tau)})=0$. So finally
we can integrate to find that for $x>\norm{\mu^{\ast}(\tau)}$
\begin{align*}
g_{\tau}^{\prime}(x) & =g_{\tau}^{\prime}(\norm{\mu^{\ast}(\tau)})+\intop_{\norm{\mu^{\ast}(\tau)}}^{x}g_{\tau}^{\prime\prime}(w)\textrm{ d} w\\
 & \leq0+\intop_{\norm{\mu^{\ast}(\tau)}}^{x}1\textrm{ d} w\\
 & =x-\norm{\mu^{\ast}(\tau)}
\end{align*}

and so integrating again gives the desired inequality:
\begin{align*}
g_{\tau}(x) & =g_{\tau}\left(\norm{\mu^{\ast}(\tau)}\right)+\intop_{\norm{\mu^{\ast}(\tau)}}^{x}g_{\tau}^{\prime}(w)\textrm{ d} w\\
 & \leq g_{\tau}\left(\norm{\mu^{\ast}(\tau)}\right)+\intop_{\norm{\mu^{\ast}(\tau)}}^{x}(w-\norm{\mu^{\ast}(\tau)})\textrm{ d} w\\
 & =g_{\tau}\left(\norm{\mu^{\ast}(\tau)}\right)+\frac{1}{2}(x-\norm{\mu^{\ast}(\tau)})^{2}
\end{align*}

The same inequality holds for $x<\norm{\mu^{\ast}(\tau)}$ by integrating
from $\intop_{x}^{\norm{\mu^{\ast}(\tau)}}.$ The final inequality
for $f$ follows by setting $x=\norm{\mu}.$
\end{proof}

\section{Aggregated Posterior}
\label{appd}

\begin{figure}[ht]
\begin{center}
\centerline{\includegraphics[scale=0.66]{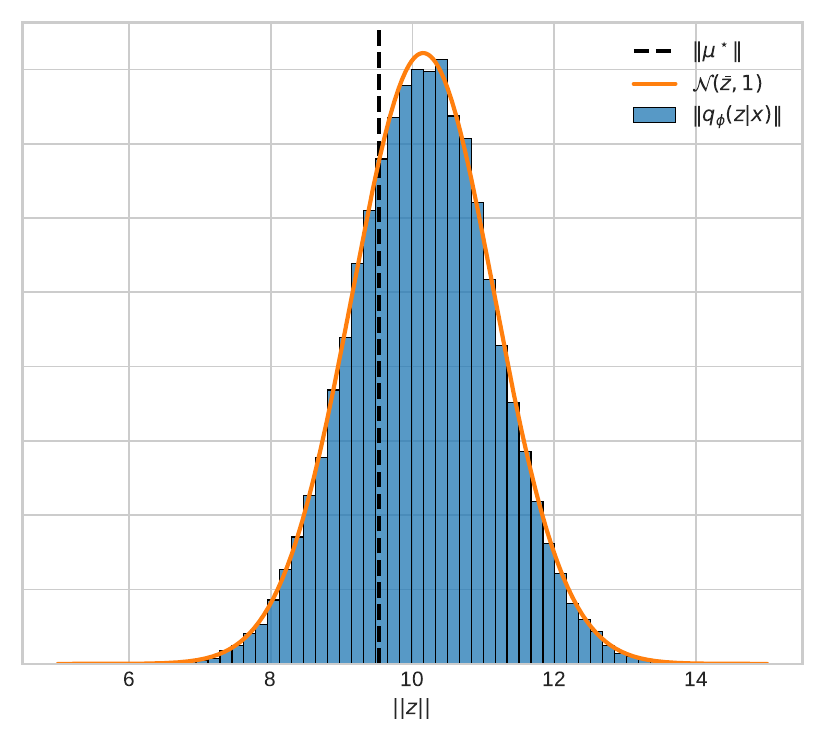}}
\caption{A comparison between the cumulative distribution function of $\mathcal{N}\left(\bar{z},1\right)$ and the radial component of $q_{\phi}\left(z \vert x\right)$. This test was performed on the FMINST dataset with $d_z=10$ and $\tau=10$.}
\label{distribution2}
\end{center}
\end{figure}

In this section we show that the aggregated posterior has a radial component which is distributed according to $\mathcal{N}\left(\bar{z},1\right)$ and provide information about the typical value of $\bar{z}$. Figure~\ref{distribution2} shows the cumulative distribution function of both $q_{\phi}\left(z \vert x\right)$ and $\mathcal{N}\left(\bar{z},1\right)$ which are approximately identical. Table~\ref{FMINIST-CIFAR-table} shows examples of $\Vert\mu^\star(\tau)\Vert$, $\bar{z}$, and $\sigma$ from models trained from the Fashion-MNIST and CIFAR-10 dataset.

\begin{table}[H]
\caption{Example values of $\bar{z}$, $\Vert\mu^\star(\tau)\Vert$ and $\sigma$ on the Fashion-MNIST and CIFAR-10 datasets}
\label{FMINIST-CIFAR-table}
\vskip 0.15in
\begin{center}
\begin{small}
\begin{sc}
\begin{tabular}{lccccr}
\toprule
Data set & $\bar{z}$ & $\sigma$ & $\Vert\mu^\star(\tau)\Vert$ & $\tau$ & $d_z$\\
\midrule
Fashion-MNIST  & 10.15 & 0.99 & 9.53 & 10  & 10\\
        & 19.91 & 1.01 & 19.77 & 20  & 10\\
        & 29.99 & 1.02 & 29.85 & 30 & 10\\
CIFAR-10 & 15.05 & 1.01 & 11.20 & 15  & 100\\
        & 24.92 & 0.97 & 22.93 & 25  & 100\\
        & 39.80 & 1.04 & 38.72 & 40  & 100\\

\bottomrule
\end{tabular}
\end{sc}
\end{small}
\end{center}
\vskip -0.1in
\end{table}

\section{Experimental Settings}
\label{appe}
\subsection{Datasets}
Datasets that are used in our experiments are MNIST, Fashion-MNIST, KMNIST, CIFAR-10, SVHN, CelebA and LSUN. We also use two synthetic datasets which we call Noise and Constant. Images from the Noise dataset are created by sampling from the uniform distribution in the range of $[0,255]$ for each pixel in an image, where the constant dataset is created by sampling from the uniform distribution in the range of $[0,255]$ where all pixels in the image have the same value. All images are resized to shape 32 x 32. Color images are converted to gray-scale for the Fashion-MNIST test by taking the first channel and discarding the other two. Grayscale images are converted to color images by taking 3 copies of the first channel. When testing with the CelebA and LSUN datasets, we use 50000 random samples from each due to the large dataset sizes.

\subsection{Model Structure and Optimization Procedure}
For all tests we train the VAE for 250 epochs with a batch size of 64.  We use the ADAM optimizer with a learning rate of $10^{-4}$ and clip gradients that are greater than 100. The encoder of the VAE consists of 5 convolutions layers with a fully connected layer for both $\mu$ and $\sigma$. The decoder uses 6 deconvolutional layers and ends with a single convolutional layer. We initialize all convolutional weights from the distribution $\mathcal{N}(0,0.2)$.  
On the Fashion-MNIST test we use latent dimensions $d_z=10$ and for the CIFAR-10 test we used $d_z=100$. Both Likelihood Regret and Likelihood Ratio are intended to be run with categorical cross entropy loss, thus we employ this as the reconstruction loss function for all comparison models except the standard Gaussian VAE in Figure~\ref{cover}, where we use the $l_2$ loss all models.

We emphasize the importance of the fully connected encoder layers for the tilted prior. This is essential as the KLD is a function of $\Vert\mu\Vert$ which is impractical to optimize with a fully convolutional model.

\subsection{Implementing Methods}
For all comparison methods we use IWAE with 200 samples to derive a lower bound on the log-likelihood. For Likelihood Regret we use gradient descent on all model parameters for 100 steps in each of the tests. We use the ADAM optimizer for this process at a learning rate of 1e-4, which is the same as the training procedure. We train the background model in Likelihood Ratio by setting the perturbation ratio parameter $\mu$ to be 0.2 for all tests. Besides this, the background model is trained in an identical format to the regular neural networks. For input complexity we use the OpenCV implementations of the PNG and JP2 compression algorithms. The tilted prior uses $\tau = 30$ for the Fashion-MNIST test and $\tau = 25$ for the CIFAR-10 test.

\section{FP80 Results}
\label{appf}

\begin{table*}[ht]
\caption{FPR-80 comparison between OOD detection methods with Fashion-MNIST and CIFAR-10 as training distributions. A smaller number is better. Results here confirm the AUCROC results in Table~\ref{AUROC-table}}
\vskip 0.15in
\begin{center}
\begin{small}
\begin{sc}
\begin{tabular}{lccccc|cc}
\toprule
Dataset & LL & IC (png) & IC (JPEG2000) & Ratio & Regret & Tilt \\
\midrule
\underline{Fashion-MNIST}    \\
MNIST           & 0.830 & 0.009 & 0.818 & 0.05 & 0.002 & \bf{0.0} \\
CIFAR-10        & \bf{0.0} & 0.034 & \bf{0.0} & 0.107 & 0.012 & 0.001 \\
SVHN            & \bf{0.0} & 0.002 & \bf{0.0} & 0.420 & 0.012 & 0.018 \\
KMNIST          & 0.512 & 0.258 & 0.494 & 0.036 & 0.002 & \bf{0.0} \\
Noise           & \bf{0.0} & 0.711 & \bf{0.0} & \bf{0.0} & 0.002 & \bf{0.0} \\
Constant        & 0.033 & \bf{0.0} & 0.025 & \bf{0.0} & \bf{0.0} & 0.543 \\
\hline
\\
\underline{CIFAR-10}       \\
MNIST           & 1.0 & 0.026 & 1.0 & 1.0 & \bf{0.020} & 0.395 & \\
Fashion-MNIST   & 1.0 & \bf{0.018} & 0.999 & 0.998 & 0.036 & 0.546 & \\
SVHN            & 0.954 & \bf{0.075} & 0.959 & 0.441 & 0.124 & 0.993 & \\
LSUN            & 0.328 & 0.977 & 0.292 & 0.713 & 0.668 & \bf{0.106} & \\
CelebA          & 0.580 & 0.963 & 0.586 & 0.851 & 0.483 & \bf{0.244} & \\
Noise           & \bf{0.0} & 0.965 & \bf{0.0} & 0.158 & 0.007 & \bf{0.0} & \\
Constant        & 1.0 & \bf{0.0} & 0.269 & 0.010 & 0.008 & 1.0 & \\

\bottomrule
\end{tabular}
\end{sc}
\end{small}
\end{center}
\vskip -0.1in
\end{table*}

\section{Reconstructions and Samples}
\begin{figure}[H]
\begin{center}
\subfigure[]{\includegraphics[width=200pt]{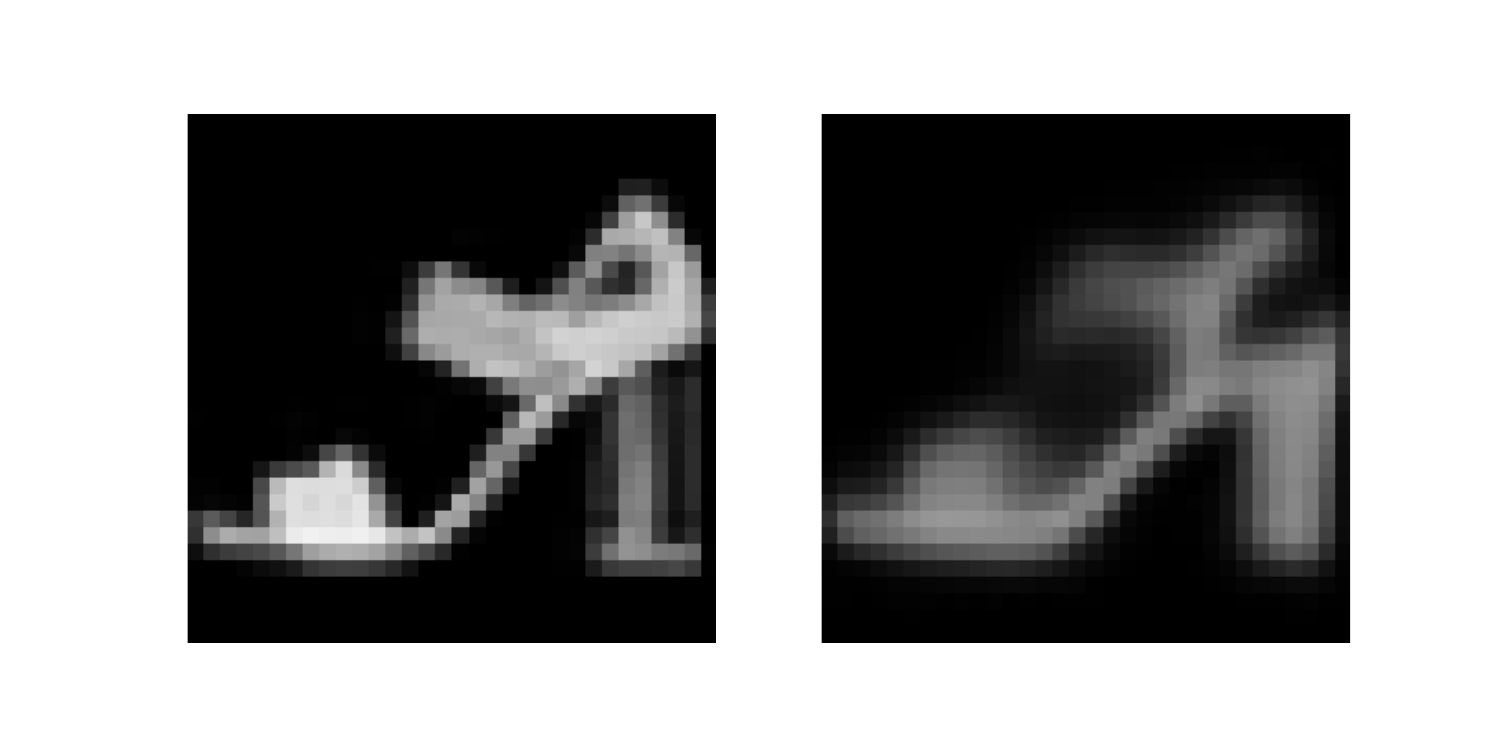}}
\subfigure[]{\includegraphics[width=200pt]{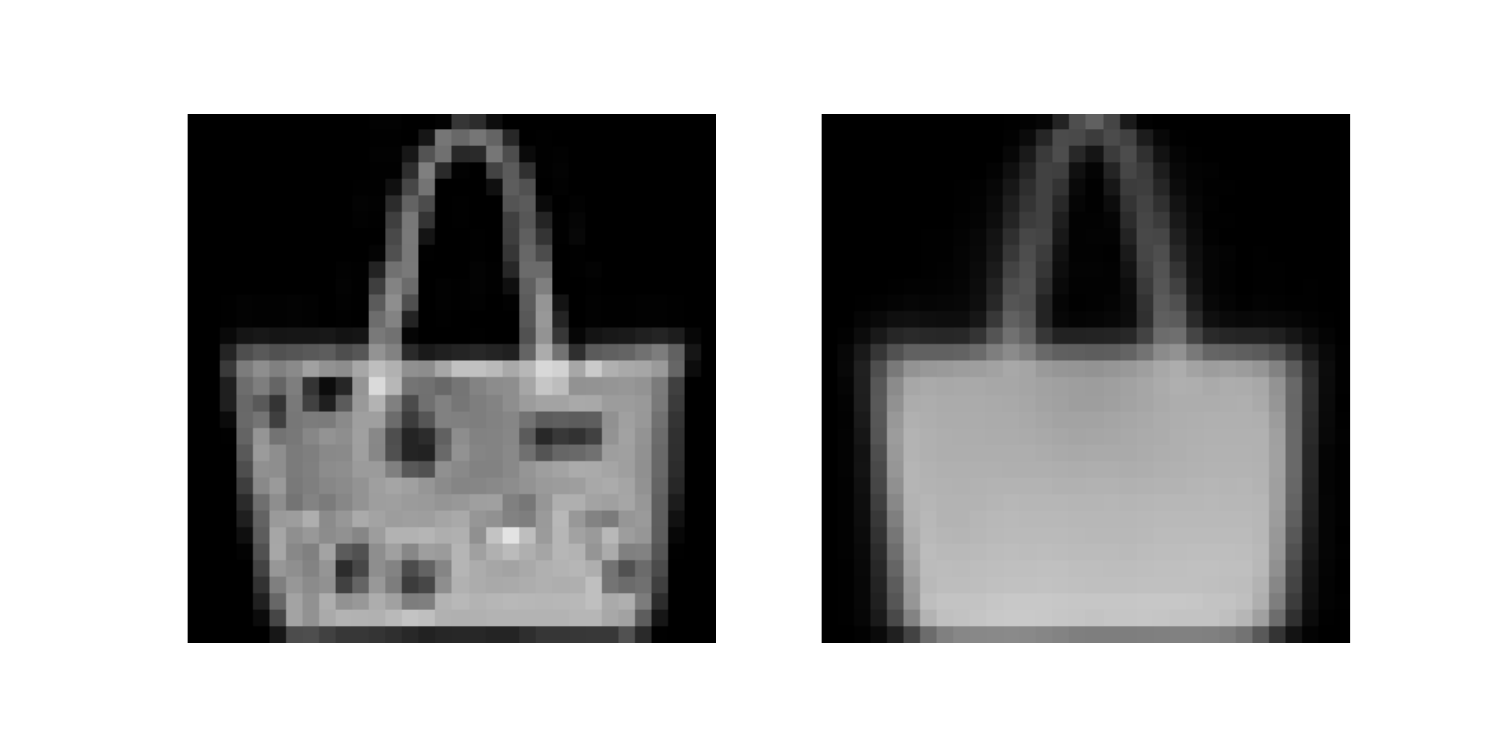}}
\subfigure[]{\includegraphics[width=200pt]{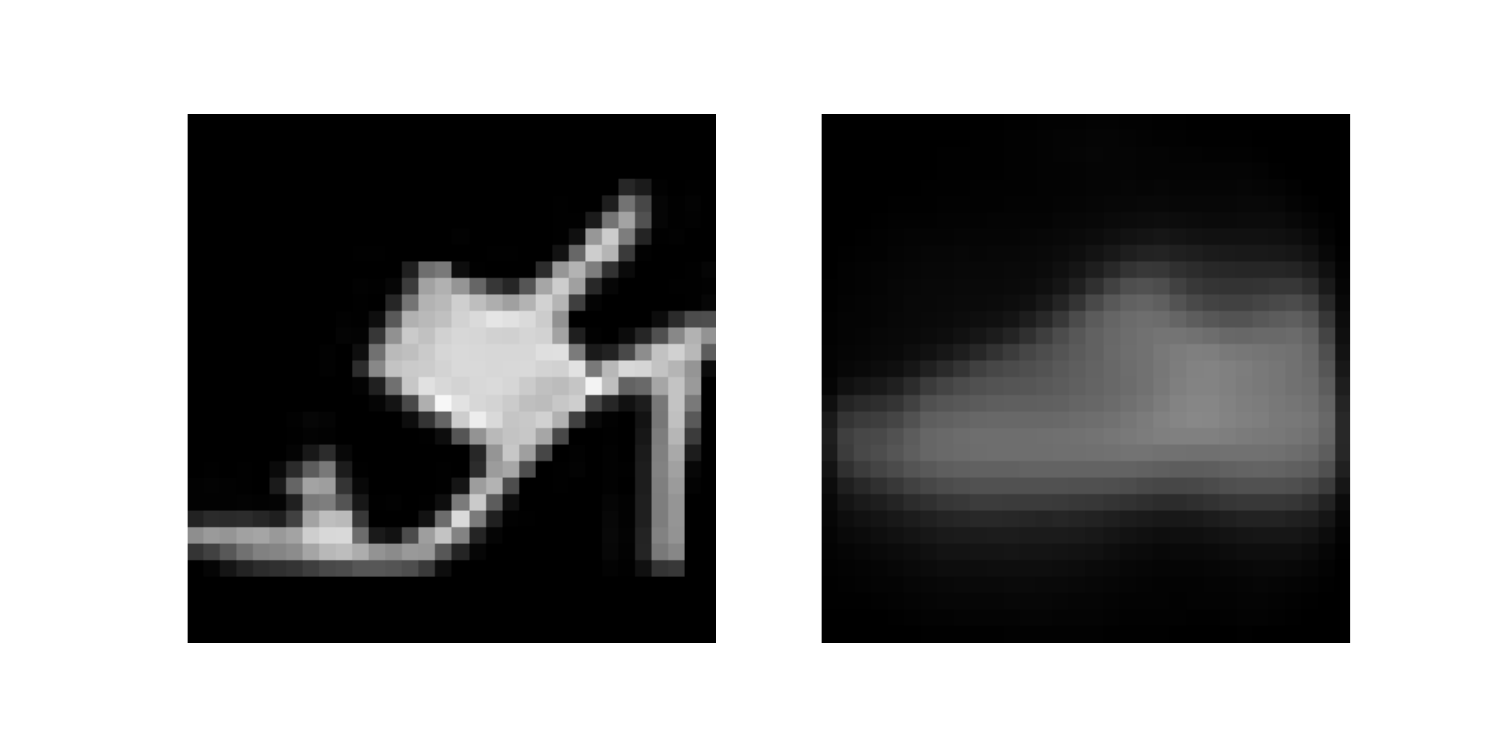}}
\subfigure[]{\includegraphics[width=200pt]{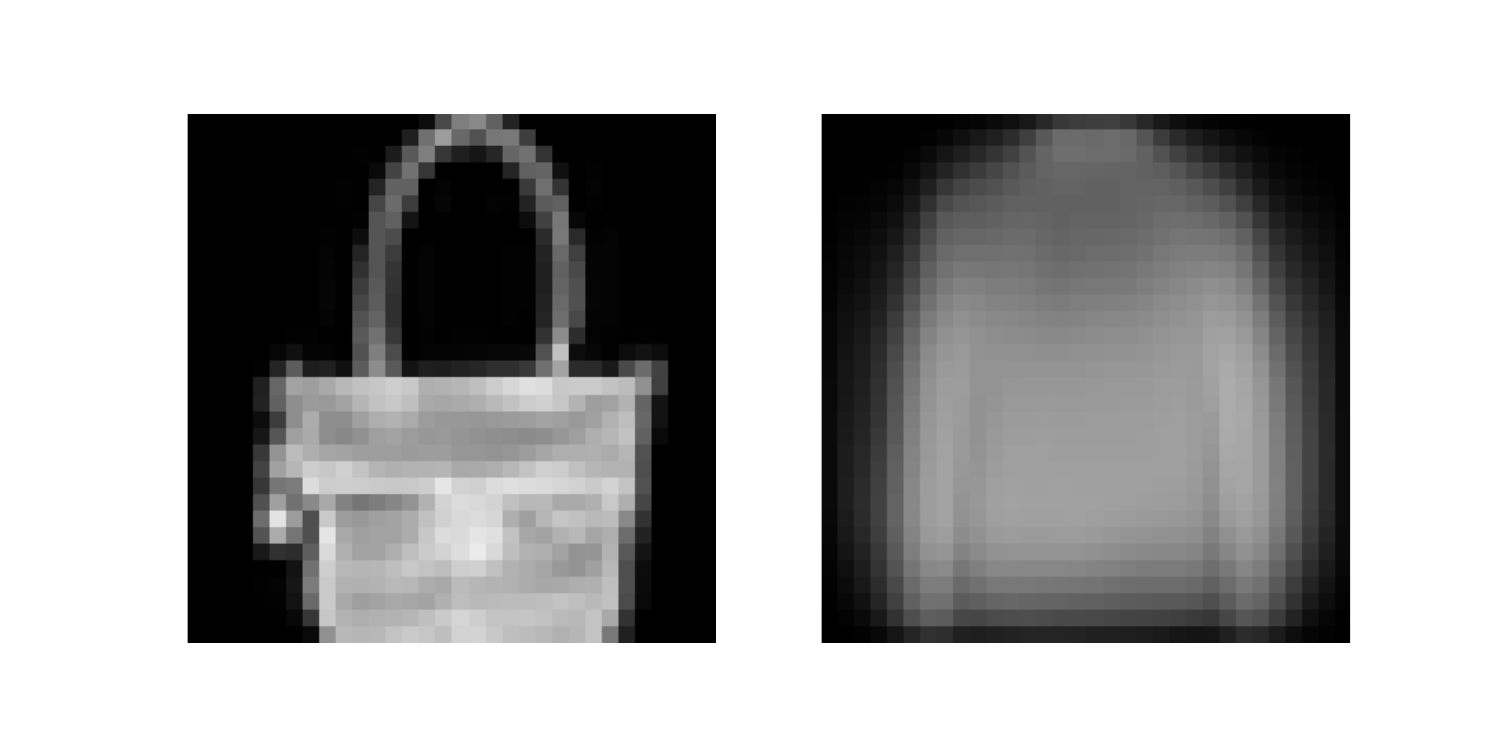}}
\caption{Comparison of reconstructed samples from the Fashion-MNIST dataset between the tilted prior and the standard Gaussian prior. Pairs of images are organized so that the left image is a sample from the original dataset and the right image is the reconstruction. Images (a) and (b) in the top row are from a VAE with a tilted prior, images (c) and (d) in the bottom row are from a VAE with a standard Gaussian prior.}
\label{MNIST-recon}

\subfigure[]{\includegraphics[width=100pt]{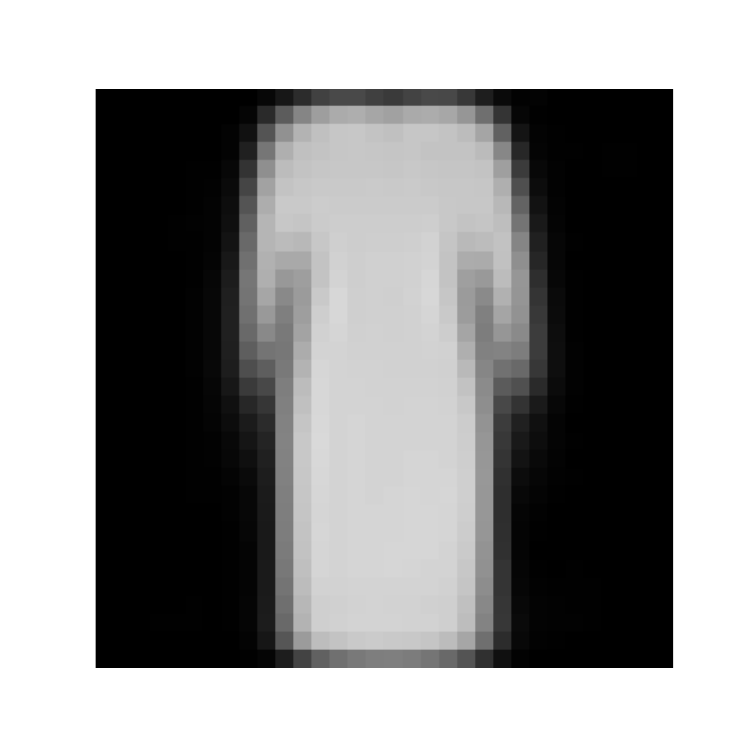}}
\subfigure[]{\includegraphics[width=100pt]{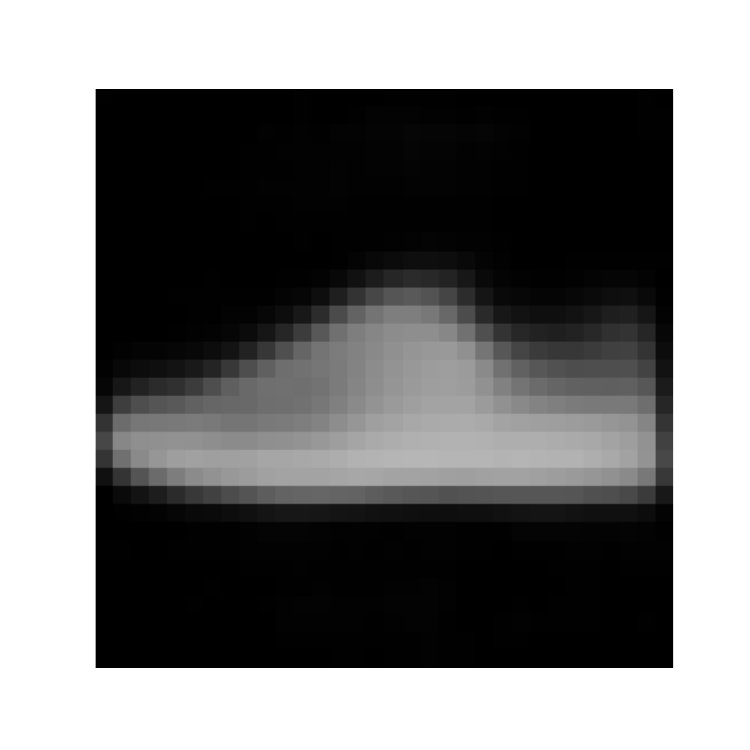}}
\subfigure[]{\includegraphics[width=100pt]{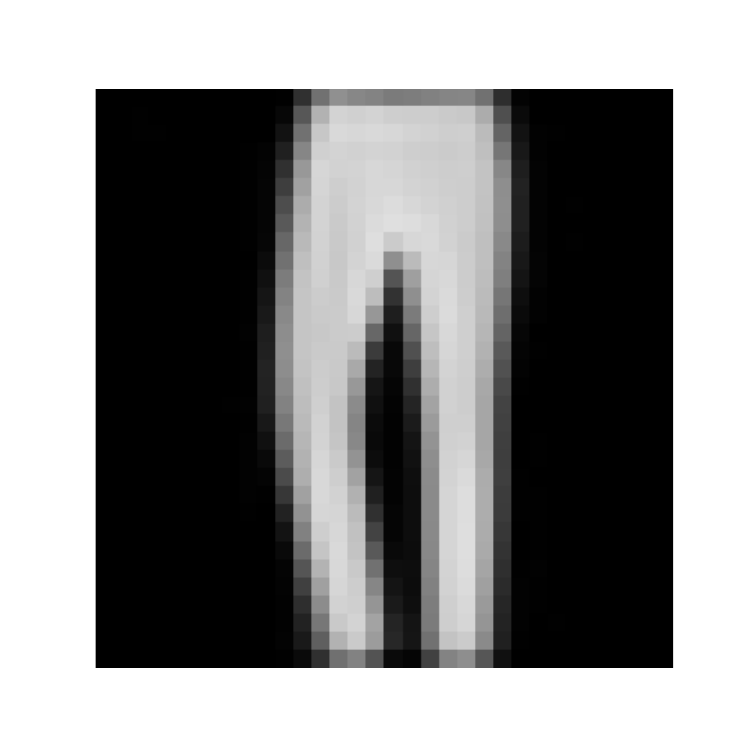}}
\subfigure[]{\includegraphics[width=100pt]{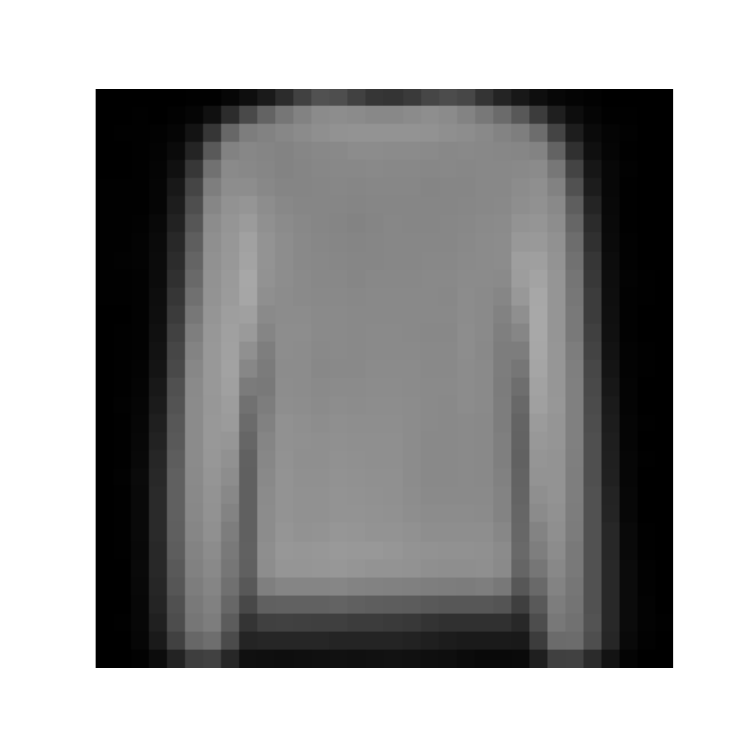}}
\subfigure[]{\includegraphics[width=100pt]{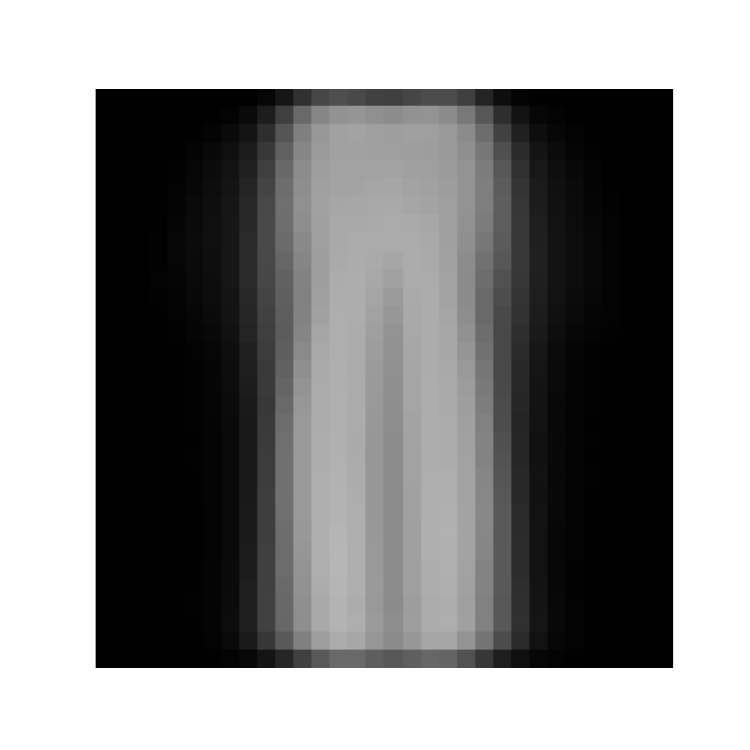}}
\subfigure[]{\includegraphics[width=100pt]{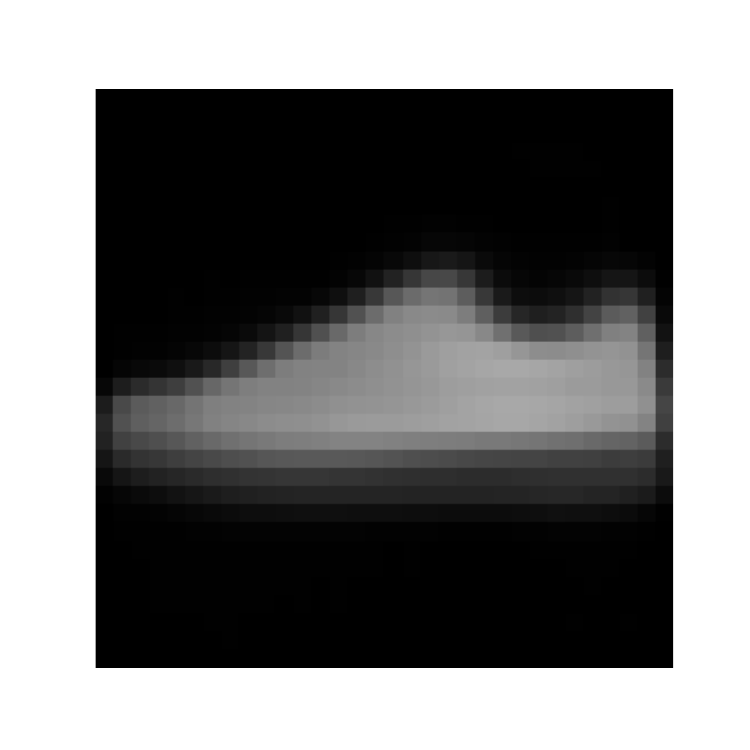}}
\subfigure[]{\includegraphics[width=100pt]{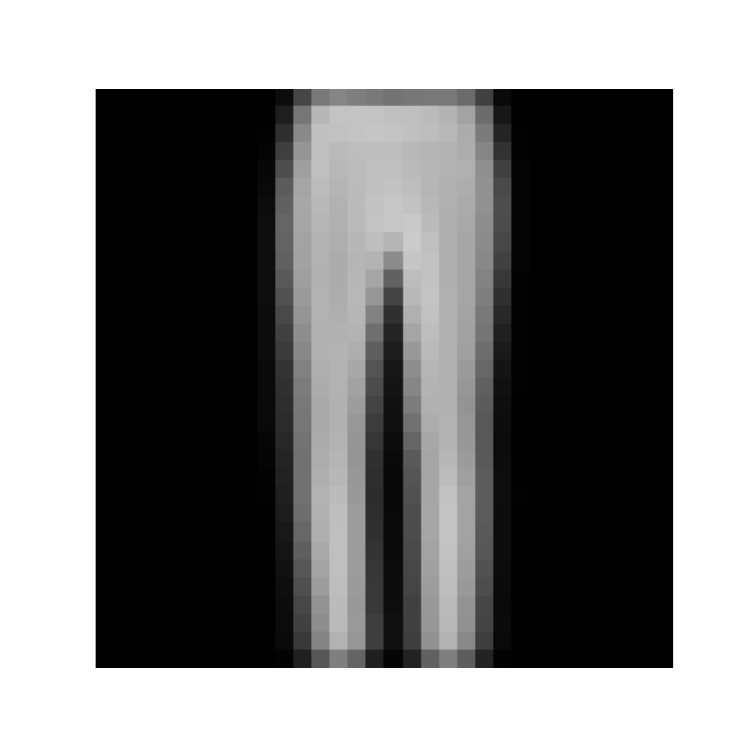}}
\subfigure[]{\includegraphics[width=100pt]{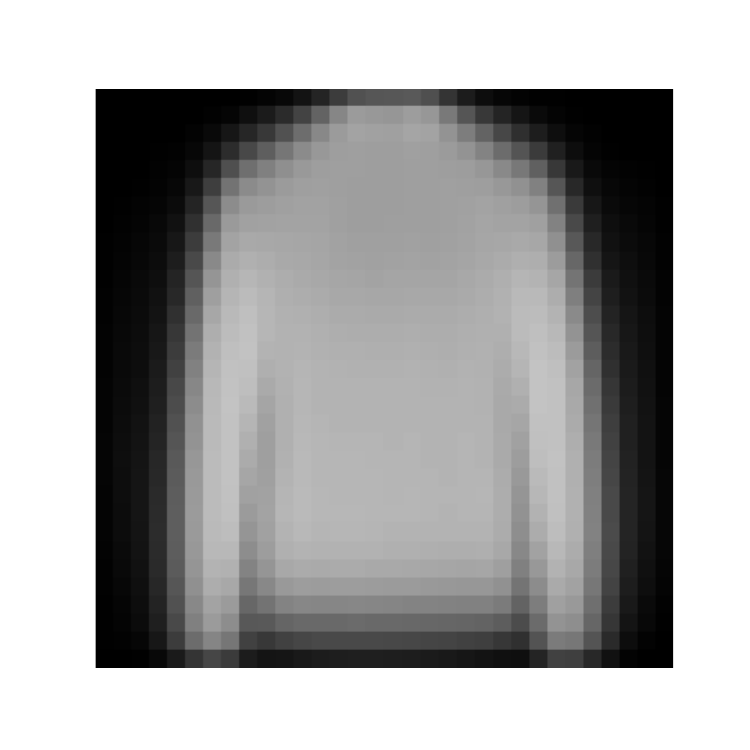}}
\caption{Comparison of image samples from models trained on the Fashion-MNIST dataset. Images (a)-(d) are from the tilted prior, images (e)-(h) are from the standard Gaussian prior. We observe that the tilted prior results in sample quality that is comparable to that from a standard Gaussian prior.}
\label{MNIST-sample}
\end{center}
\end{figure}

\end{document}